%% file: main.tex
\documentclass[10pt,twocolumn,letterpaper]{article}

\usepackage{cvpr} 
\usepackage{comment}

\input{preamble}

\definecolor{cvprblue}{rgb}{0.21,0.49,0.74}
\usepackage[pagebackref,breaklinks,colorlinks,allcolors=cvprblue]{hyperref}

\def\paperID{5619} 
\def\confName{CVPR}
\def\confYear{2025}

\title{\networkname: Mapping using Transformers for Volumes -- Network \\for Super-Resolution with Long-Range Interactions}

\author{
August Leander Høeg, Sophia W. Bardenfleth, Hans Martin Kjer, Tim B. Dyrby,\\ Vedrana Andersen Dahl, Anders Dahl\\
Visual Computing, Technical University of Denmark\\
{\tt\small \{aulho, soeba, hmkj, tbdy, vand, abda\}@dtu.dk}
}

\begin{document}
\maketitle
\input{sec/00_abstract}

\input{sec/01_ny_intro_August}

\input{sec/02_related}

\input{sec/03_methods}

\input{sec/04_experimental_setup}

\input{sec/05_implementation_details}

\input{sec/06_results}

\input{sec/07_ablation_study}

\input{sec/08_conclusion}

{\small
\bibliographystyle{ieeenat_fullname}
\bibliography{main}
}

\input{sec/A1_suppl}

\end{document}

%% file: preamble.tex
\newcommand{\todo}[1]{{\color{red}#1}}

\newcommand{\networkname}{MTVNet\xspace}

\usepackage{tikz} 
\usepackage{subcaption} 
\usepackage{nicematrix} 
\usepackage[hang,flushmargin]{footmisc}

\usepackage{pifont}
\newcommand{\cmark}{\ding{51}}%
\newcommand{\xmark}{\ding{55}}%

\newcommand{\floor}[1]{\lfloor #1 \rfloor} 

%% file: sec/00_abstract.tex
\begin{abstract}
\noindent Until now, it has been difficult for volumetric super-resolution to utilize the recent advances in transformer-based models seen in 2D super-resolution. The memory required for self-attention in 3D volumes limits the receptive field. Therefore, long-range interactions are not used in 3D to the extent done in 2D and the strength of transformers is not realized. We propose a multi-scale transformer-based model based on hierarchical attention blocks combined with carrier tokens at multiple scales to overcome this. Here information from larger regions at coarse resolution is sequentially carried on to finer-resolution regions to predict the super-resolved image. Using transformer layers at each resolution, our coarse-to-fine modeling limits the number of tokens at each scale and enables attention over larger regions than what has previously been possible. We experimentally compare our method, \networkname, against state-of-the-art volumetric super-resolution models on five 3D datasets demonstrating the advantage of an increased receptive field. This advantage is especially pronounced for images that are larger than what is seen in popularly used 3D datasets. Our code is available at \url{https://github.com/AugustHoeg/MTVNet}.
\end{abstract}

%% file: sec/01_ny_intro_August.tex
\section{Introduction}
\label{sec:intro}

In recent years, super-resolution (SR) and other vision tasks have seen significant improvements via usage of vision transformers (ViTs). Although ViTs achieve state-of-the-art (SOTA) performance in 2D SR \cite{Liang_SwinIR, Chen_Activating, Hsu_DRCT, Chu_HMANet}, few studies have attempted applying ViTs for volumetric SR. Part of the success of ViTs is their increased receptive field compared to Convolutional Neural Networks (CNNs), enabling inferences based on broader image context \cite{Dosovitskiy_ViT_16x16words}. 
In volumetric SR, ViTs are challenged by the cubic growth in tokens required to process larger 3D image contexts. Although window-based attention improves the quadratic complexity of attention mechanisms \cite{Liu_SwinTransformer}, the complexity of 3D data still limits the receptive field of volumetric ViT-based models. Because of this disadvantage, the performance gap of CNNs vs.\ transformer-based architectures for volumetric SR has yet to be fully understood. 

Several works have studied visual enhancement of 3D medical data such as MRI (magnetic resonance imaging) and CT (computed tomography) by upscaling each slice independently \cite{oktay_2DcardiacSR, Wang_accelerating, Song_fetalSR, Xia_BrainSR, Jiang_CT_SR}. While such approaches circumvent the complexity issues of volumetric SR, not fully considering the 3D context sacrifices performance and risks inter-slice discontinuities \cite{Chen_mDCSRN_A, Chen_mDCSRN_B, Chen_mDCSRN_C, Pham_ReCNN, Forigua_SuperFormer}.



Current brain MRI benchmark datasets for evaluating volumetric SR are relatively low-resolution \cite{Ji_3DSR_survey}, limiting the benefits of a larger receptive field. Advancements in medical imaging technology enable higher spatial resolution \citep{wehrse2023ultrahigh}, resulting in larger volumes where volumetric SR can benefit from long-range contextual information. Given the potential of SR in clinical settings and the increasing interest in applications like multi-resolution synchrotron imaging \citep{walsh2021imaging}, there is a need for volumetric SR methods designed specifically for high-resolution (HR) 3D data.

\begin{figure}[t]
   \centering
   \includegraphics[width=1.0\linewidth]{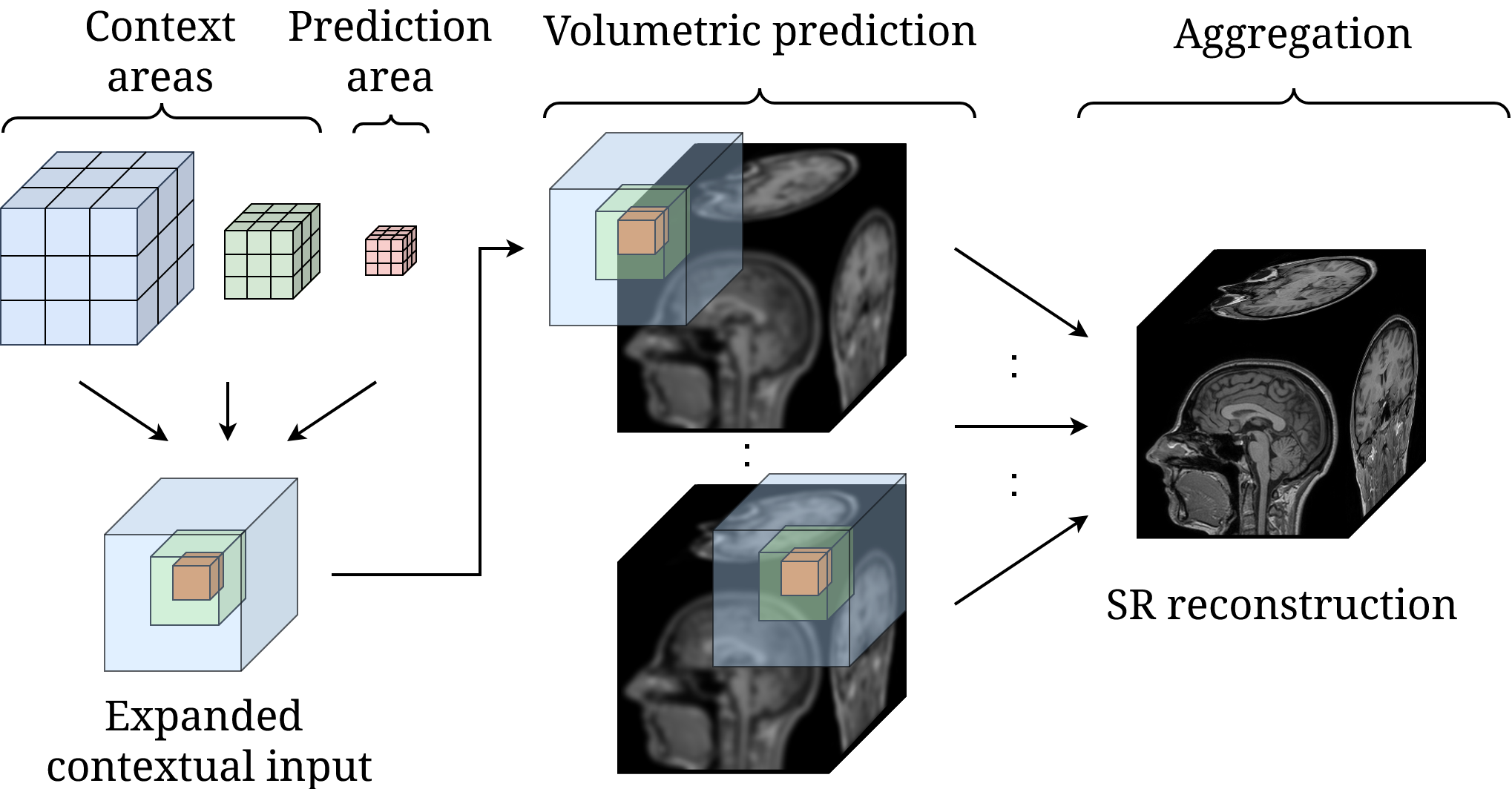}
   \caption{Overview of \networkname that is informed by a large contextual volume processed at multiple resolution scales for predicting SR in the center volume.}
   \label{fig:overview_of_network}
\end{figure}

Aside increasing contextual information in volumetric SR, recent studies in 2D SR have shown that the window-based attention mechanism of the Swin-Transformer \cite{Liu_SwinTransformer} is not ideal for capturing relationships across distant image regions. Using Local Attribution Mapping (LAM), \citet{Chen_Activating} showed that strengthening long-range information exchange can lead to significant performance gains. Similarly, recent studies in ViT architectures have focused on modeling long-range interactions to increase performance \citep{Chen_CrossViT, Hatamizadeh_FasterViT}. 

To address these limitations, we present \networkname, a volumetric SR approach based on multi-scale image representation and hierarchical attention to enhance long-range information propagation. Our \networkname broadens the receptive field by expanding the contextual input beyond the prediction area, see \cref{fig:overview_of_network}. We hypothesize that for the SR task, image regions near the prediction region provide the most important contextual information while more distant regions still provide relevant information, but contribute less. Consequently, we design a coarse-to-fine feature extraction and tokenization scheme with progressively less computational resources allocated towards regions further from the prediction area, enabling us to increase the volumetric input size without exceeding GPU memory. 
Furthermore, inspired by FasterViT \cite{Hatamizadeh_FasterViT} and SwinV2 \cite{Liu_SwinV2}, we propose an efficient shifting hierarchical attention mechanism suitable for volumetric image processing. This approach leverages specialized carrier tokens (CATs) that contain compact feature summaries of larger attention windows. Using full attention in the highly compressed CAT domain, our model improves modeling of long-range spatial information, further improving SR performance in volumetric data. 




We compare our proposed \networkname against several volumetric SR approaches on four brain MRI benchmark datasets and one high-resolution CT based dataset. Extensive experiments show that convolutional models still outperform ViT-based architectures in lower resolution datasets. Although on high-resolution 3D data with meaningful long-range image dependencies, our proposed \networkname outperforms all other volumetric SR approaches. We anticipate that our proposed multi-contextual approach could greatly benefit other volumetric image tasks.  

%% file: sec/02_related.tex
\section{Related Work}
\label{sec:related}

\subsection{Learning-based super-resolution}
The advantages of learning-based SR over classical interpolation methods were first demonstrated by SRCNN proposed by \citet{Dong_SRCNN}. Since then, several CNN-based SR models have been proposed to improve performance and computational efficiency \citep{Dong_FSRCNN,Ledig_SRGAN,Lim_EDSR,Wang_ESRGAN,Zhang_RCAN}. Despite the success of CNNs, many vision tasks such as image classification \cite{Dosovitskiy_ViT_16x16words, Liu_SwinTransformer, Chen_CrossViT, Hatamizadeh_FasterViT}, object detection \cite{Carion_DETR, Tahira_ObjectDetection_Review, Gao_AdaMixer, Roh_SparseDETR}, segmentation \cite{Cao_Swin_Unet, Chen_TransUNet, Wang_TransBTS, Hatamizadeh_SwinUNETR, Gao_UTNetV2}, and SR have seen improvements using vision transformers (ViTs).  
SwinIR by \citet{Liang_SwinIR} were among the first to demonstrate the superiority of transformers over convolution-based models for SR by incorporating the Swin Transformer \cite{Liu_SwinTransformer} in a residual network scheme. \citet{Chen_HAT,Chen_Activating} proposed cross attention of overlapping window partitions and channel attention mechanisms to enable activation of more input pixels. \citet{Chu_HMANet} suggested HMANet, which integrates a grid-shuffling scheme with window-based attention to model cross-area similarity for enhanced image reconstruction. Very recently, \citet{Hsu_DRCT} have suggested combining Swin-transformer layers and gating mechanisms in a densely-connected structure \citep{Huang_DenseNet,Tong_SRDenseNet} to alleviate information bottlenecks. 

Concurrently, improvements to the vision transformer backbone have been proposed to enable efficient processing of HR image data. \citet{Liu_SwinV2} proposed SwinV2, featuring improved normalization and a more efficient attention mechanism using cosine similarly. This work was later applied to SR by \citet{Conde_Swin2SR} in Swin2SR. In CrossViT \citep{Chen_CrossViT}, multi-scale tokenization and efficient cross-attention mechanisms were used to extract and fuse feature representations at different image scales. Recently, \citet{Hatamizadeh_FasterViT} proposed FasterViT, an efficient vision transformer including local window attention and global attention. Since these models focus on 2D images, most do not scale well in 3D, requiring substantial modifications to be applied for volumetric data.  

\begin{figure*}[t]
  \centering
    \includegraphics[width=0.90\linewidth]{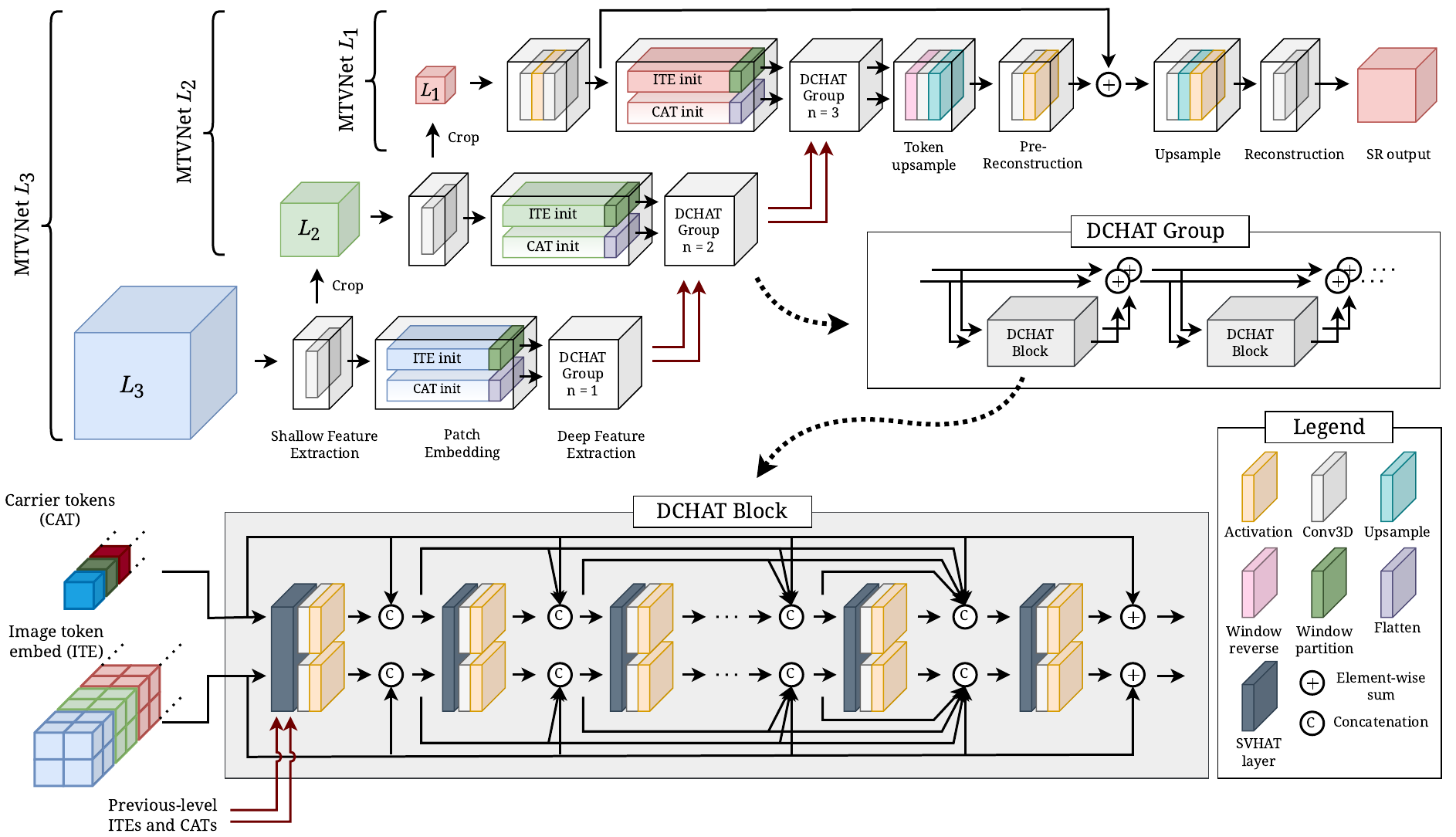}
    \caption{Illustration of \networkname and the structure of DCHAT Block and DCHAT Group. Our proposed architecture consists of up to three levels of multi-contextual volumetric image processing. 
    The first two levels perform tokenization using larger 3D patch sizes to cover broader contextual regions, while succeeding levels process subsets of the input volume using smaller patch sizes, resulting in both coarse- and fine-grained feature extraction. The depth of subsequent DCHAT Groups increases from $n = 1$ to $3$ DCHAT Blocks towards the last stage. The token embeddings from preceding network levels are fused into later levels using cross attention.}
    \label{fig:network_arch}
\end{figure*}

\subsection{Super-resolution for 3D volumes}
3D SR methods operate slice-wise or volumetric. Slice-wise methods predict each slice independently and typically leverage model architectures from 2D SR. While these models can handle entire slices simultaneously and often support deeper network architectures, they lack cross-slice information, which can lead to discontinuities between slice predictions.

Volumetric SR methods fully utilize the context in 3D, achieving better overall performance than slice-wise methods because of improved inter-plane predictions, but with much higher computational costs \cite{Chen_mDCSRN_A, Chen_mDCSRN_B, Chen_mDCSRN_C, Pham_ReCNN, Forigua_SuperFormer}. Inspired by SRCNN \citep{Dong_SRCNN} and SRGAN \citep{Chen_mDCSRN_A}, \citet{Pham_SRCNN3D} and \citet{Chen_mDCSRN_A} proposed three-dimensional adaptations of convolutional SR models and demonstrated the potential of volumetric SR over slice-wise approaches. Research in volumetric SR has since grown rapidly and several methods have been proposed to improve efficiency and performance \citep{Chen_mDCSRN_B,Chen_mDCSRN_C,Du_DCED,Li_MFER,Lu_novel,Wu_ArSSR,Pham_ReCNN,Sanchez_SRGAN3D,Wang_EDDSR}. These approaches are very similar to classical SR in that they aim to predict HR reconstructions from isotropically degraded images, only on volumetric instead of 2D images. However, several other approaches for volumetric SR also exist. For instance, to account for the fact that clinical MR images often feature high in-plane and low through-plane resolution, axial SR models \citep{Ge_ResVoxGAN,Huang_TransMRSR,Wang_ASFT} have been proposed to increase slice count while preserving in-plane resolution. 

To alleviate the limitation of fixed upscaling factors, arbitrary scale SR based on Implicit Neural Representation \cite{Wu_ArSSR} \cite{Zhu_MIASSR} \cite{Li_McASSR} have been proposed. Another branch of volumetric SR is multi-contrast models \citep{Li_McASSR, Huang_TransMRSR} that leverage information from multiple MRI modalities (T1- and T2-weighted images).

Transformer-based architectures have also been proposed for volumes. SuperFormer \citep{Forigua_SuperFormer} merged feature embeddings and volume embeddings using a volumetric transformer-based network structure similar to SwinIR \citep{Liang_SwinIR}. Also inspired by SwinIR, \citet{Huang_TransMRSR} implemented a transformer-based GAN (generative adversarial network) model for axial super-resolution using residual swin transformer blocks \citep{Liang_SwinIR,Liu_SwinTransformer}. The CFTN model \citep{Zhang_CFTN} employed 3D residual channel attention blocks \citep{Zhang_RCAN} and transformers to capture global cross-scale dependencies between multi-scale feature embeddings. 
\citet{Li_McASSR} proposed a 2D slice-wise multi-modal arbitrary scale SR model featuring a rectangle-window cross-attention transformer to model longer-range image dependencies.
Despite the growing interest in volumetric transformer-based SR models, several of the improvements seen in 2D SR cannot be effectively applied in 3D due to memory limitations. Our work seeks to leverage these developments to enhance volumetric SR while simultaneously addressing the challenge of limited contextual information, a critical bottleneck in volumetric SR.

%% file: sec/03_methods.tex
\section{Methods}
\label{sec:methods}

\begin{figure*}[t]
  \centering
  \begin{subfigure}{0.132\linewidth} 
    \includegraphics[width=1.0\linewidth]{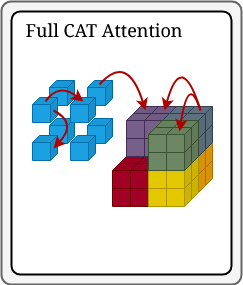}
    \caption{CAT attention.}
    \label{fig:full_cat_attn}
  \end{subfigure}
  \hfill
  \begin{subfigure}{0.385\linewidth} 
    \includegraphics[width=1.0\linewidth]{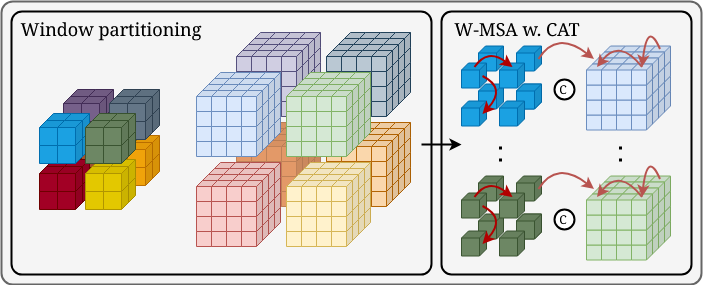}
    \caption{Window partitioning followed by W-MSA w. CAT.}
    \label{fig:msa_w_cat}
  \end{subfigure}
  \hfill
  \begin{subfigure}{0.473\linewidth} 
    \includegraphics[width=1.0\linewidth]{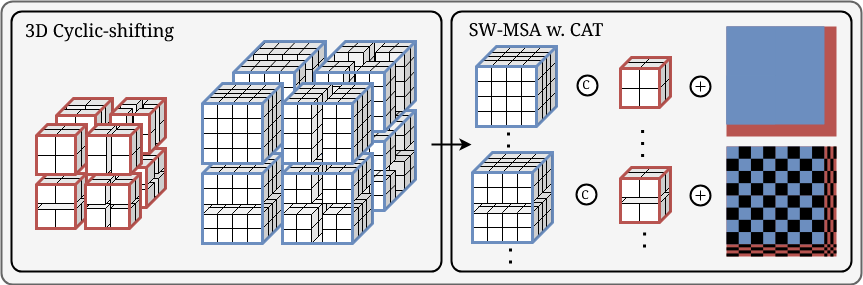}
    \caption{3D Cyclic-shifting followed by SW-MSA w. CAT}
    \label{fig:swmsa_w_cat}
  \end{subfigure}
  \caption{Illustration of volumetric attention mechanisms used in SVHAT: \ref{fig:full_cat_attn}) Full CAT attention, \ref{fig:msa_w_cat}) W-MSA with CAT and \ref{fig:swmsa_w_cat}) SW-MSA with CAT. Our proposed SVHAT uses alternating shifted and non-shifted windowed attention. Masking is used to limit information exchange between non-adjacent ITEs and CATs. In these examples, the window size is $M = 4$ and the CAT space size is $c=2$.}
  \label{fig:masked_attention}
\end{figure*}

\subsection{Network architecture}
The architecture of \networkname consists of three levels of volumetric image processing: $L_{3}$, $L_{2}$, and $L_{1}$, see \cref{fig:network_arch}. The network levels $L_{3}$ and $L_{2}$ extract features from image regions surrounding the SR prediction area and merges these features into $L_{1}$. These features serve as a prior for the network level $L_{1}$, enabling conditioning of the SR output based on the surrounding image context.

\textbf{Shallow Feature Extraction}. Our \networkname uses shallow feature extraction (SFE) modules for initial processing at each level. Given an input volume $I_{\text{LR}} \in \mathbb{R}^{C_{\text{in}} \times H \times W \times D}$, each SFE module expands the channel dimension using $3\times3\times3$ convolutional layers, producing shallow features $\mathcal{F}_{\text{SFE}} \in \mathbb{R}^{C_{\text{emb}} \times H \times W \times D}$. The output from each stage's SFE is cropped and passed to the next level, allowing the model to leverage the features of previous SFE modules.

\textbf{Patch embedding}.  
During patch embedding, shallow features are projected and tokenized into differently-sized volumetric image patches. The levels $L_{3}$ and $L_{2}$ use larger patch sizes to cover wider image regions, reducing the number of tokens required for processing these volumes. Image token embeddings (ITEs) $\mathbf{x}^{L}$ are obtained via a $p_{{L}}\times p_{{L}}\times p_{{L}}$ strided convolution, where $L$ corresponds to the network level. For subsequent processing, we partition the ITEs into attention windows of $M \times M \times M$ tokens. The corresponding carrier token embeddings (CATs) $\mathbf{x}_{\text{cat}}^{L}$ are initialized from the ITEs using convolution with stride and kernel size $\floor{\frac{M}{c}} \times \floor{\frac{M}{c}} \times \floor{\frac{M}{c}}$, where $c$ is a factor determining the number of CATs for each attention window. 

\textbf{Deep Feature Extraction}.
Deep feature extraction is performed within each level using DCHAT (densely connected hierarchical attention) blocks to extract high-frequency spatial information. The DCHAT blocks for each level are connected in a residual scheme to produce DCHAT groups consisting of up to three DCHAT blocks. In the case of multiple levels, cross-attention mechanisms \citep{Vaswani_Transformer} are used to merge token embeddings into subsequent network levels, facilitating the propagation of multi-scale information.

\textbf{Reconstruction}.
In the final stage, token upsampling is performed via a deconvolution layer, transforming the patch embeddings back into the image space. These features are further refined in a pre-reconstruction stage before being fused with the shallow features through a long skip-connection. The fused features are then upsampled using a 3D pixel-shuffle layer \cite{Shi_pixelshuffle}. We employ a 3D pre-convolution layer initialized according to the ICNR method described in \cite{Aitken_Checkerboard_artifact_free} to prevent checkerboard artifacts during pixel-shuffling.

\subsection{Dense-Connected Hierarchical Attention block}
For efficient extraction of volumetric image features, we propose a DCHAT block, see \cref{fig:network_arch}. Inspired by \citet{Hsu_DRCT}, our DCHAT block employs a densely connected structure of volumetric transformer layers, LeakyReLU activations, and convolutions. To preserve the feature space of ITEs and CATs, we process each set of tokens using separate sets of skip connections and convolutions. Additionally, we match the embedding dimension of ITEs and CATs throughout each block to equally promote learning of progressively complex features. As in DRCT \citep{Hsu_DRCT} we utilize $1 \times 1 \times 1$ convolutions as gating mechanisms between transformer layers to filter redundant features, improving efficiency and enabling feature transition between DCHAT blocks.

\subsection{Shifting Hierarchical Attention Transformer}
Inspired by FasterViT \citep{Hatamizadeh_FasterViT} and SwinV2 \cite{Liu_SwinV2}, we design an SVHAT (shifting volumetric hierarchical attention transformer) layer for concurrent processing of ITEs and CATs. Similar to FasterViT, SVHAT uses a combination of full attention and windowed attention to extract hierarchical image features. The attention mechanisms used in SVHAT are illustrated in \cref{fig:masked_attention}. First, full attention in the CAT space allows global information flow across attention windows, see \cref{fig:full_cat_attn}. 
Next, we concatenate each attention window's corresponding CATs and ITEs, providing each attention window access to its set of CATs.   
Windowed attention is then applied jointly to the ITEs and CATs to capture token dependencies, with the CATs conveying global information from other attention windows, see \cref{fig:msa_w_cat}. This alternating attention procedure allows global feature exchange between local attention windows, improving long-range information flow. To further enhance information exchange, we reintroduce shifted window-based attention into the attention framework proposed in FasterViT \citep{Hatamizadeh_FasterViT}, see \cref{fig:swmsa_w_cat}. Before window partitioning of ITEs and CATs, we perform 3D cyclic-shifting to allow attention of tokens in neighboring windows. To account for the presence of CATs, we shift both the image space and CAT space by $\floor{\frac{M}{2}}$ and $\floor{\frac{c}{2}}$ voxels, respectively. This shifting conserves the alignment of the two feature spaces. Attention masking is applied to drop interactions between non-adjacent tokens in the ITE/CAT space.

We compute attended carrier token embeddings $\mathbf{x}_{\text{cat}}^{L, t}$ at network level $L$ and transformer layer $t$ as follows:
\begin{equation}
    \begin{split}
        \hat{\mathbf{x}}_{\text{cat}}^{L, t} & = {\mathbf{x}}_{\text{cat}}^{L, t-1} +\gamma_1 \operatorname{MSA}\left(\operatorname{LN}\left({\mathbf{x}}_{\text{cat}}^{L, t-1} \right)\right), \\
        {\mathbf{x}}_{\text{cat}}^{L, t} & = \hat{\mathbf{x}}_{\text{cat}}^{L, t} + \gamma_2 \operatorname{MLP}\left(\operatorname{LN}\left(\hat{\mathbf{x}}_{\text{cat}}^{L, t} \right)\right),
    \end{split}
    \label{eq:ct_attention}
\end{equation}

\noindent where $\gamma_{1}$, $\gamma_{2}$ are learnable channel-wise scaling factors, $\operatorname{MSA}$ is the multi-headed self-attention mechanism \cite{Vaswani_Transformer}, $ \operatorname{LN}$ denotes Layer Normalization \cite{Ba_LayerNorm}, and $\operatorname{MLP}$ is the multi-layer perception.

After CAT attention, we compute attention of ITEs and CATs using windowed self-attention, see \cref{eq:x_ct_attention}. The CATs are window partitioned and concatenated with their corresponding set of ITEs to produce sequences of $M^{3} + c^{3}$ tokens for each attention window. 
Inspired by SwinV2 \citep{Liu_SwinV2}, SVHAT employs a post-normalized Shifted Window based Self-Attention (SW-MSA) procedure. Window-attended CATs and ITEs ${\mathbf{x}}_{\mathbf{w}}^{L,t+1}$ are computed as:
\begin{equation}
    \begin{split}
        \mathbf{x}_{\mathbf{w}}^{L, t} & = [ {\mathbf{x}}^{L, t-1}, \hspace{0.1cm} {\mathbf{x}}_{\text{cat}}^{L,t} ] \\
        \hat{\mathbf{x}}_{\mathbf{w}}^{L,t+1} & = {\mathbf{x}}_{\mathbf{w}}^{L,t} + \operatorname{LN} \left(\operatorname{SW-MSA}\left({\mathbf{x}}_{\mathbf{w}}^{L, t}\right)\right) \\
        {\mathbf{x}}_{\mathbf{w}}^{L,t+1} & = \hat{\mathbf{x}}_{\mathbf{w}}^{L,t+1} + \operatorname{LN} \left(\operatorname{MLP} \left( \hat{\mathbf{x}}_{\mathbf{w}}^{L,t+1} \right) \right)
    \end{split}
    \label{eq:x_ct_attention}
\end{equation}
The ITEs and CATs are then separated again to ensure compatibility with subsequent SVHAT layers. 

Prior to the attention mechanisms for ITEs and CATs (described in \cref{eq:ct_attention} and \cref{eq:x_ct_attention}), SVHAT uses multi-head cross-attention (MCA) layers to facilitate information exchange across network levels. Each cross-attention layer implements a two-layer MLP to ensure dimension compatibility between cross-scale token sequences. Then, MCA is applied to capture relationships between current- and previous-level token embeddings. Exploiting the small size of the CAT space, we compute cross-attended CATs $\mathbf{x}^{L}_{\text{cross, cat}}$ using full MCA between all current-level and previous-level CATs:
\begin{equation}
        \mathbf{x}^{L}_{\text{cross, cat}} = \operatorname{LN}\left(\operatorname{MCA}\left(\mathbf{x}_{\text{cat}}^{L, t-1}, \operatorname{MLP}\left(\mathbf{x}_{\text{cat}}^{L-1}\right)\right)\right),
    \label{eq:ct_cross_attention}
\end{equation}
where $\mathbf{x}_{\text{cat}}^{L-1}$ denotes the final set of CATs from the previous network level. A similar window-based multi-head cross-attention (W-MCA) mechanism is used for capturing relationships between current- and previous-level ITEs, see equation \ref{eq:x_cross_attention}. The cross-attended ITEs $\mathbf{x}^{L}_{\text{cross}}$ are computed as follows:
\begin{equation}
        \mathbf{x}^{L}_{\text{cross}} = \operatorname{LN}\left(\operatorname{W-MCA}\left(\mathbf{x}^{L, t-1}, \operatorname{MLP}\left(\mathbf{x}^{L-1}\right)\right)\right),
    \label{eq:x_cross_attention}
\end{equation}
\noindent where $\mathbf{x}^{L-1}$ denote the final set of ITEs from the previous network level. Finally, the cross-attended token embeddings are fused using a residual scheme:
\begin{equation}
    \begin{split}
        {\mathbf{x}}_{\text{cat}}^{L, t-1} & = \bar{\mathbf{x}}_{\text{cat}}^{L, t-1} + \mathbf{x}^{L}_{\text{cross, cat}} \\
        {\mathbf{x}}^{L, t-1} & = \bar{\mathbf{x}}^{L, t-1} + \mathbf{x}^{L}_{\text{cross}}
    \end{split}
    \label{eq:x_ct_cross_attention_merge}
\end{equation}
Here, $\bar{\mathbf{x}}^{L, t-1}$ and $\bar{\mathbf{x}}_{\text{cat}}^{L, t-1}$ denote ITEs and CATs before fusion. To reduce the complexity of \networkname, cross-attention is performed only in the first SVHAT layer of every DCHAT block. 

%% file: sec/04_experimental_setup.tex
\begin{table*}[t]
\centering
\resizebox{\textwidth}{!}{

\renewcommand{\arraystretch}{1.}

\begin{NiceTabular}{l!{\!\!}c!{\!\!\!\!}c!{\!\!\!\!}c!{\quad}c!{\!\!\!\!}c!{\!\!\!\!}c!{\qquad}c!{\!\!\!\!}c!{\!\!\!\!}c!{\quad}c!{\!\!\!\!}c!{\!\!\!\!}c}
\toprule
& \Block{1-6}{FACTS-Synth Dataset} & & & & & & \Block{1-6}{FACTS-Real Dataset} \\
& \Block{1-3}{Scale $4 \times$} & & & \Block{1-3}{Scale $3 \times$} & & & \Block{1-3}{Scale $4 \times$} & & & \Block{1-3}{Scale $3 \times$} \\
Method & \footnotesize{PSNR} & \footnotesize{SSIM} & \footnotesize{NRMSE} & \footnotesize{PSNR} & \footnotesize{SSIM} & \footnotesize{NRMSE} & \footnotesize{PSNR} & \footnotesize{SSIM} & \footnotesize{NRMSE} & \footnotesize{PSNR} & \footnotesize{SSIM} & \footnotesize{NRMSE}\\
\midrule
ArSSR & $28.83$ & $.8998$ & $.1779 $ & $30.78$ & $.9284$ & $.1459$ & $20.88$ & $.3871$ & $.4881$ & $20.68$ & $.3980$ & $.5767$ \\ 

EDDSR & $29.86$ & $.9109$ & $.1620$ & $33.22$ & $.9451$ & $.1104$ & $20.62$ & $.3531$ & $.4815$ & $19.84$ & $.3499$ & $.5223$\\ 

MFER & $29.48$ & $.9094$ & $.1646$ & $32.50$ & $.9420$ & $.1179$ & $\textcolor{blue}{21.58}$ & $\textcolor{red}{.4708}$ & $.4080$ & $21.64$ & $\textcolor{blue}{.4671}$ & $.4096$ \\ 

mDCSRN & $29.77$ & $.9099$ & $.1624$ & $33.23$ & $.9460$ & $.1090$ & $21.31$ & $.4078$ & $.4765$ & $21.37$ & $.4259$ & $.4922$\\ 

SuperFormer & $\textcolor{blue}{30.46}$ & $\textcolor{blue}{.9175 }$ & $\textcolor{blue}{.1481}$ & $\textcolor{blue}{33.47}$ & $\textcolor{blue}{.9480}$ & $\textcolor{blue}{.1055}$  & $\textcolor{black}{20.93}$ & $\textcolor{black}{.3491}$ & $\textcolor{black}{.4846}$ & $\textcolor{black}{21.40}$ & $\textcolor{black}{.4038}$ & $\textcolor{black}{.4463}$ \\ 

RRDBNet3D & $\textcolor{black}{29.78}$ & $\textcolor{black}{.9120}$ & $\textcolor{black}{.1584}$ & $33.21$ & $.9442$ & $.1093$ & $\textcolor{red}{21.64}$ & $\textcolor{blue}{.4670}$ & $\textcolor{red}{.4022}$ & $\textcolor{red}{21.91}$ & $\textcolor{red}{.4775}$ & $\textcolor{red}{.4019}$ \\ 

MTVNet & $\textcolor{red}{31.57}$ & $\textcolor{red}{.9303}$ & $\textcolor{red}{.1313}$ & $\textcolor{red}{33.91}$ & $\textcolor{red}{.9502}$ & $\textcolor{red}{.1020}$ & $\textcolor{black}{21.52}$ & $\textcolor{black}{.4576}$ & $\textcolor{blue}{.4061}$ & $\textcolor{blue}{21.74}$ & $\textcolor{black}{.4633}$ & $\textcolor{blue}{.4051}$\\ 
\bottomrule
\toprule

& \Block{1-6}{HCP 1200 Dataset} & & & & & & \Block{1-6}{IXI Dataset} \\
& \Block{1-3}{Scale $4 \times$} & & & \Block{1-3}{Scale $2 \times$} & & & \Block{1-3}{Scale $4 \times$} & & & \Block{1-3}{Scale $2 \times$} \\
Method & \footnotesize{PSNR} & \footnotesize{SSIM} & \footnotesize{NRMSE} & \footnotesize{PSNR} & \footnotesize{SSIM} & \footnotesize{NRMSE} & \footnotesize{PSNR} & \footnotesize{SSIM} & \footnotesize{NRMSE} & \footnotesize{PSNR} & \footnotesize{SSIM} & \footnotesize{NRMSE}\\
\midrule
ArSSR & $27.90$ & $.8118$ & $.2810$ & $35.54$ & $.9372$ & $.1212$ &  $24.22$ & $.7204$ & $.3060$ & $30.32$ & $.9152$ & $.1560$\\ 

EDDSR & $30.12$ & $.8335$ & $.2174$ & $35.19$ & $.9317$ & $.1274$ & $25.22$ & $.7394$ & $.2597$ & $33.45$ & $.9447$ & $.1101$ \\ 

MFER & $33.40$ & $.8933$ & $.1484$ & $37.24$ & $.9498$ & $.1011$ & $25.23$ & $.7611$ & $.2576$ & $35.67$ & $.9622$ & $.0865$\\ 

mDCSRN & $33.46$ & $.8941$ & $.1470$ & $37.18$ & $.9493$ & $.1017$ & $29.50$ & $.8558$ & $.1622$ & $35.66$ & $.9619$ & $.0863$\\ 

SuperFormer & $\textcolor{black}{33.70}$ & $\textcolor{black}{.8982 }$ & $\textcolor{black}{.1430}$ & $\textcolor{black}{36.65}$ & $\textcolor{black}{.9441}$ & $\textcolor{black}{.1080}$ & $\textcolor{black}{29.89}$ & $\textcolor{black}{.8679}$ & $\textcolor{black}{.1545}$ & $\textcolor{black}{35.00}$ & $\textcolor{black}{.9575}$ & $\textcolor{black}{.0925}$\\ 

RRDBNet3D & $\textcolor{red}{34.31}$ & $\textcolor{red}{.9092}$ & $\textcolor{red}{.1331}$ & $\textcolor{red}{37.70}$ & $\textcolor{red}{.9533}$ & $\textcolor{red}{.0959}$ & $\textcolor{red}{30.27}$ & $\textcolor{red}{.8793}$ & $\textcolor{red}{.1488}$ & $\textcolor{red}{36.92}$ & $\textcolor{red}{.9697}$ & $\textcolor{red}{.0755}$ \\ 

MTVNet & $\textcolor{blue}{34.04}$ & $\textcolor{blue}{.9046}$ & $\textcolor{blue}{.1374}$ & $\textcolor{blue}{37.53}$ & $\textcolor{blue}{.9520}$ & $\textcolor{blue}{.0978}$ & $\textcolor{blue}{30.16}$ & $\textcolor{blue}{.8754}$ & $\textcolor{blue}{.1502}$ & $\textcolor{blue}{36.38}$ & $\textcolor{blue}{.9668}$ & $\textcolor{blue}{.0799}$ \\ 
\bottomrule
\toprule

& \Block{1-6}{BraTS 2023 Dataset} & & & & & & \Block{1-6}{Kirby 21 Dataset} \\
& \Block{1-3}{Scale $4 \times$} & & & \Block{1-3}{Scale $2 \times$} & & & \Block{1-3}{Scale $4 \times$} & & & \Block{1-3}{Scale $2 \times$} \\
Method & \footnotesize{PSNR} & \footnotesize{SSIM} & \footnotesize{NRMSE} & \footnotesize{PSNR} & \footnotesize{SSIM} & \footnotesize{NRMSE} & \footnotesize{PSNR} & \footnotesize{SSIM} & \footnotesize{NRMSE} & \footnotesize{PSNR} & \footnotesize{SSIM} & \footnotesize{NRMSE}\\
\midrule
ArSSR & $22.96 $ & $.3182 $ & $.4437 $ & $36.67 $ & $.9691 $ & $.1079 $ & $31.82$ & $.8632$ & $.2775$ & $43.94$ & $.9860$ & $.0705$\\ 

EDDSR & $32.66$ & $.9169$ & $.1686$ & $38.29$ & $.9766$ & $.0916$ & $33.51$ & $.8946$ & $.2244$ & $44.42$ & $.9874$ & $.0693$\\ 

MFER & $34.76$ & $.9430$ & $.1309$ & $41.88$ & $.9867$ & $.0614$  & $35.68$ & $.9307$ & $.1719$ & $51.51$ & $.9970$ & $.0309$\\ 

mDCSRN & $34.76$ & $.9431$ & $.1308$ & $41.88$ & $.9865$ & $.0614$ & $35.26$ & $.9255$ & $.1806$ & $50.64$ & $.9962$ & $.0345$ \\ 

SuperFormer & $\textcolor{black}{34.60}$ & $\textcolor{black}{.9400 }$ & $\textcolor{black}{.1333}$ & $\textcolor{black}{40.46}$ & $\textcolor{black}{.9831}$ & $\textcolor{black}{.0716}$ & $\textcolor{black}{35.85}$ & $\textcolor{black}{.9341}$ & $\textcolor{black}{.1675}$ & $\textcolor{black}{48.14}$ & $\textcolor{black}{.9936}$ & $\textcolor{black}{.0453}$\\ 

RRDBNet3D & $\textcolor{red}{35.20}$ & $\textcolor{red}{.9486}$ & $\textcolor{red}{.1242}$ & $\textcolor{red}{43.76}$ & $\textcolor{red}{.9894}$ & $\textcolor{red}{.0501}$ & $\textcolor{red}{36.27}$ & $\textcolor{red}{.9376}$ & $\textcolor{red}{.1598}$ & $\textcolor{red}{56.36}$ & $\textcolor{red}{.9988}$ & $\textcolor{red}{.0175}$ \\ 

MTVNet & $\textcolor{blue}{35.16}$ & $\textcolor{blue}{.9477}$ & $\textcolor{blue}{.1250}$ & $\textcolor{blue}{42.71}$ & $\textcolor{blue}{.9880}$ & $\textcolor{blue}{.0560}$  & $\textcolor{blue}{35.97}$ & $\textcolor{blue}{.9355}$ & $\textcolor{blue}{.1654}$ & $\textcolor{blue}{53.60}$ & $\textcolor{blue}{.9977}$ & $\textcolor{blue}{.0247}$\\ 
\bottomrule

\end{NiceTabular}

} 

\caption{Quantitative comparison of state-of-the-art volumetric SR models on datasets FACTS-Synth, FACTS-Real, HCP 1200, IXI, BraTS 2023, and Kirby 21. The best performance metrics PSNR $\uparrow$ / SSIM $\uparrow$ / NRMSE $\downarrow$ are highlighed in \textcolor{red}{red}, and the second best in \textcolor{blue}{blue}.}
\vspace{-0.05in}
\label{tab:table_quantitative_results}
\end{table*}

\section{Experiments}
\label{sec:experimental_setup}

\subsection{Experimental setup}
\textbf{Datasets}. 
We use four public MRI datasets and one CT-based dataset to train and evaluate our proposed \networkname:
The Human Connectome Project (HCP) 1200 Subjects dataset \cite{Van_HCP_1200_dataset}, the IXI dataset\footnote{\url{https://brain-development.org/ixi-dataset/}}, the Brain Tumor Segmentation Challenge (BraTs) 2023 \cite{Baid_BraTS_GLI, Menze_BraTS_GLI, Bakas_BraTS_GLI, Bakas_BraTS2023_GLI_2} and Kirby 21 \cite{Landman_Kirby21_dataset}. These datasets consist of multi-modality image volumes acquired using 1.5T-3T MRI platforms with a volume size of $\leq 320^{3}$ voxels. 
The last dataset considered is the Femur Archaeological CT Superresolution (FACTS) dataset \cite{Bardenfleth_FACTS}, which consists of 12 registered 3D volume pairs of archaeological femur bones scanned using clinical-CT and micro-CT. The FACTS dataset features large volumes ($\sim 2000^{3}$ voxels), enabling us to showcase the benefits of additional contextual information. Two SR tasks are considered using this dataset: In FACTS-Synth, we use downsampled versions of the micro-CT images as the SR model input, while FACTS-Real instead uses the clinical-CT images. The training/test splits for all datasets will be detailed in supplementary material. 

\textbf{Models}. To demonstrate the effectiveness of our proposed method, we evaluate the performance of \networkname against six other volumetric SR models: mDCSRN \citep{Chen_mDCSRN_C}, EDDSR \citep{Wang_EDDSR}, MFER \citep{Li_MFER}, RRDBNet3D \cite{Wang_ESRGAN}, SuperFormer \citep{Forigua_SuperFormer}, and ArSSR \citep{Wu_ArSSR}. 
We modify mDCSRN and SuperFormer, which were originally designed to restore images degraded by 3D k-space truncation, a method that simulates LR MRI acquisition \citep{Chen_mDCSRN_A,Forigua_SuperFormer}, by extending these models using the same 3D pixel-shuffle upsampling module used in \networkname.
For EDDSR, MFER, RRDBNet3D and ArSSR, we use the authors' suggested upsampling approach.

\textbf{Training}. We train all models from scratch on each dataset for 100K iterations on a single NVIDIA A100 80GB GPU. For ArSSR, we collate sets of $N=8000$ randomly sampled HR/LR point pairs from 15 patches for each batch.
The remaining models use a batch size of 5 and LR patch size of $32\times32\times32$, except \networkname $L_{2}$ and $L_{3}$ where we use $64\times64\times64$ and $128\times128\times128$, respectively. All models are optimized using ADAM \citep{Kingma_ADAM} with $\beta_{1} = 0.9$ and $\beta_{2} = 0.999$. We use a multi-step learning rate scheduler, halving the learning rate once after 50k, 70k, 85k, and 95k iterations. The model parameters are optimized using a simple L1 loss metric. HR/LR pairs are generated using volumetric blurring followed by downsampling via linear interpolation. In FACTS-Real, we use the clinical-CT images as LR input but omit blurring since the LR images are already smooth.

\textbf{Evaluation}. For evaluation, we reconstruct all volumetric samples in the test set of each respective dataset using strided aggregation of SR predictions. We tile each SR prediction using an overlap of $4 \times s$ voxels where $s$ is the upscaling factor and smooth the overlapping prediction areas using a Hanning window. The performance metrics Peak-Signal-to-Noise Ratio (PSNR), Structural Similarity Index Measure (SSIM), and Normalized Root Mean Square Error (NRMSE) are computed slice-wise in the axial direction and averaged over all samples in each dataset, ignoring any slices where the foreground occupies less than 25\% of the voxels.

%% file: sec/05_implementation_details.tex
\begin{figure*}[t]
  \centering
  \begin{subfigure}{0.49\linewidth}
    \includegraphics[width=1.0\linewidth]{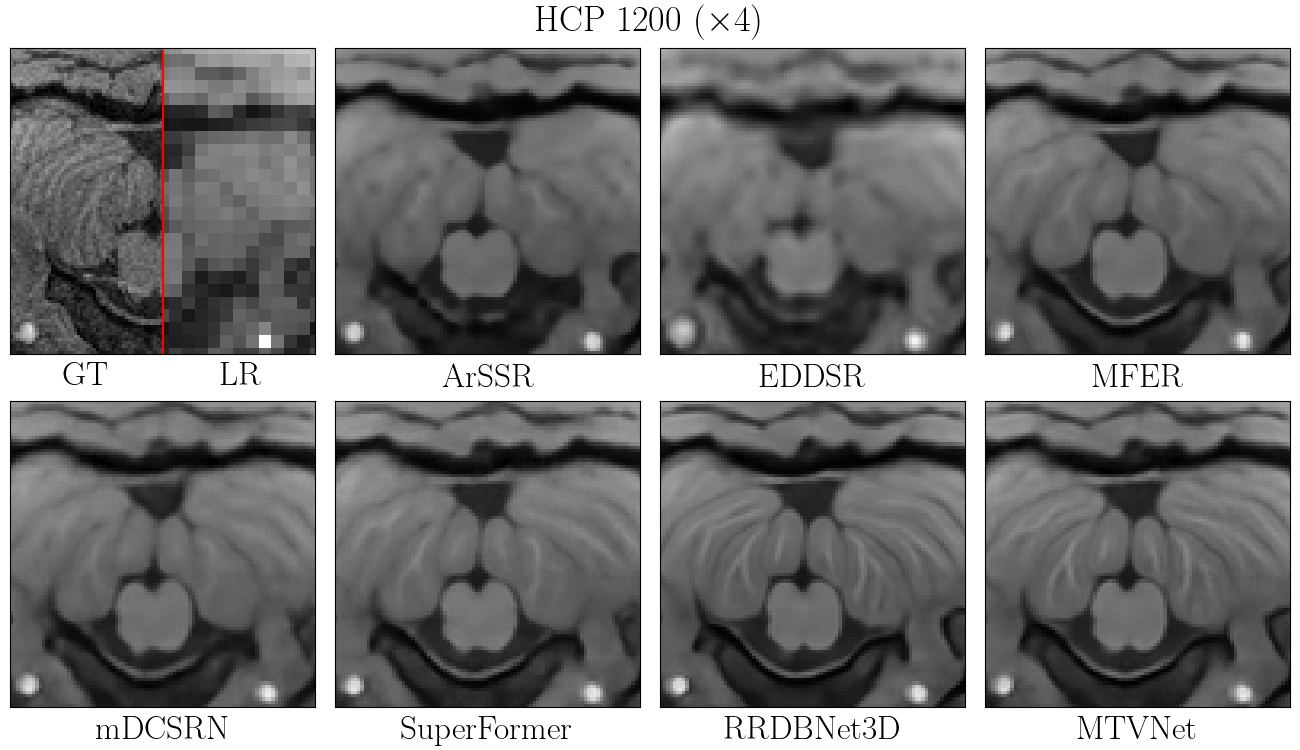}
  \end{subfigure}
  \hfill
  \begin{subfigure}{0.49\linewidth}
    \includegraphics[width=1.0\linewidth]{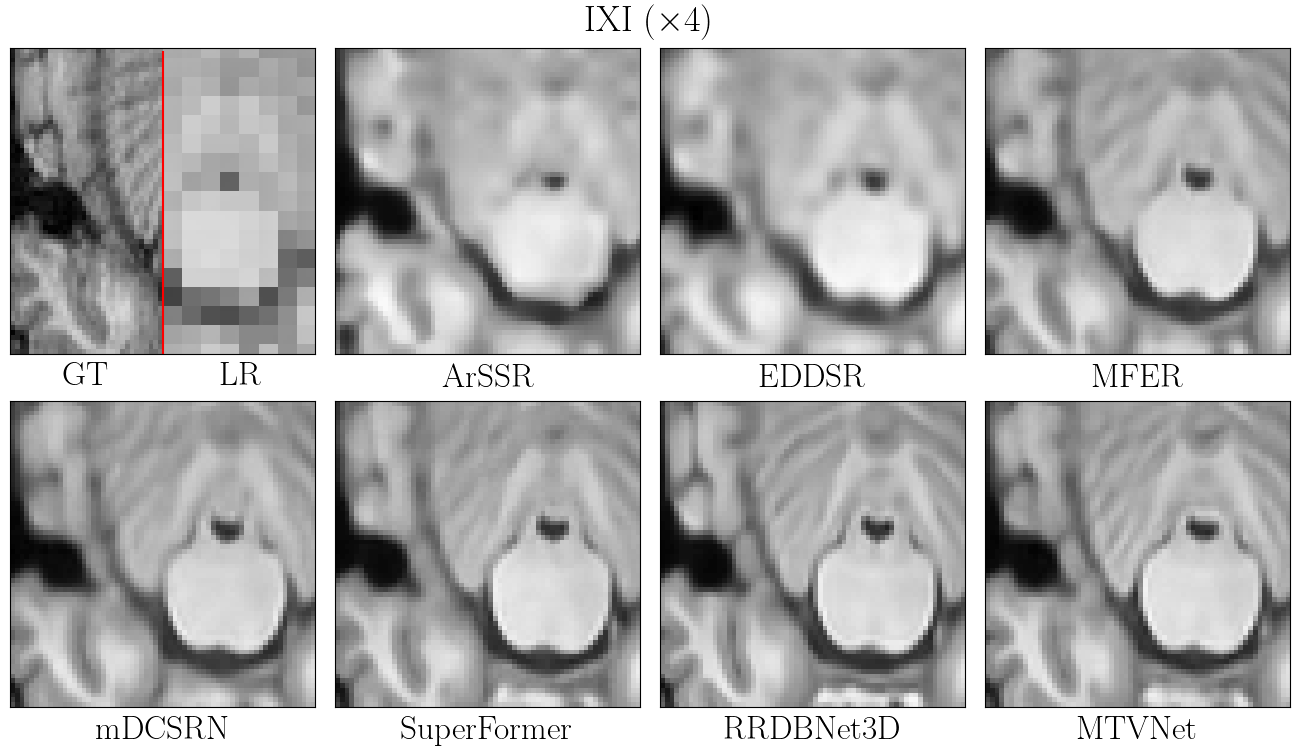}
  \end{subfigure}
  
  \vspace{0.5em} 

  \begin{subfigure}{0.49\linewidth}
    \includegraphics[width=1.0\linewidth]{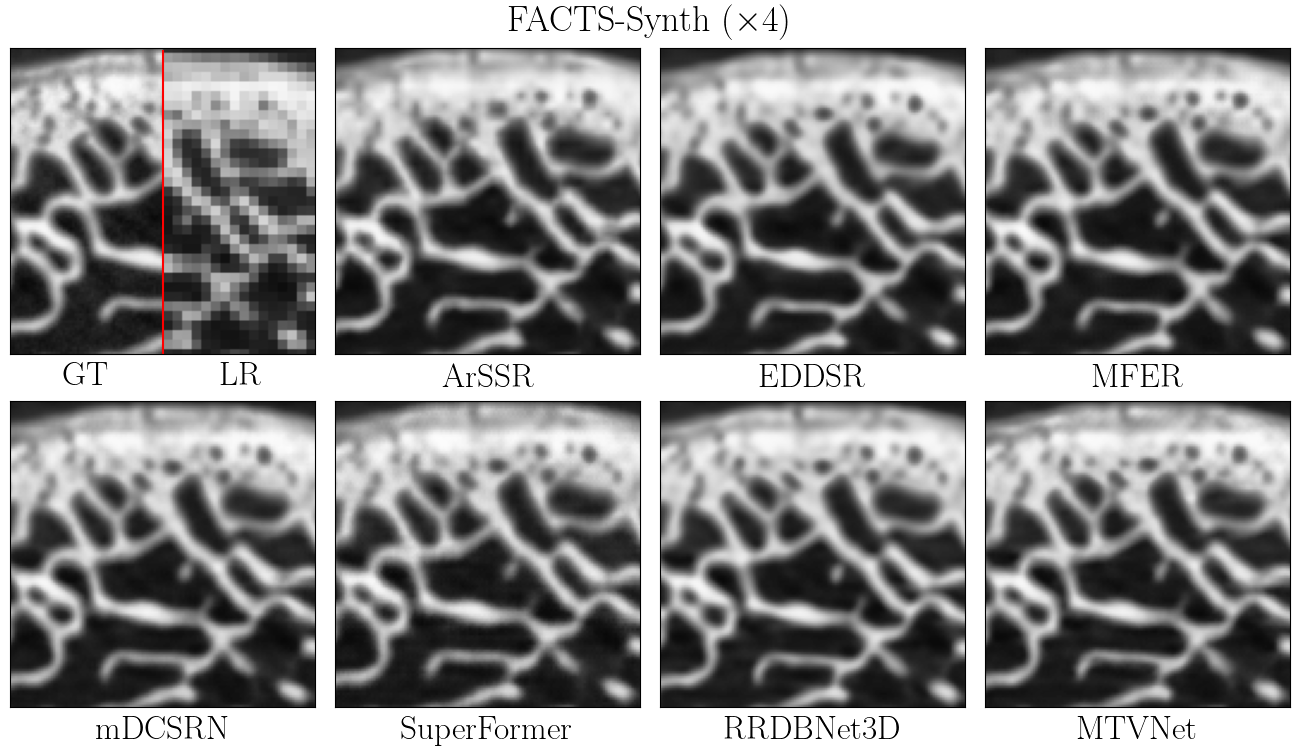}
  \end{subfigure}
  \hfill
  \begin{subfigure}{0.49\linewidth}
    \includegraphics[width=1.0\linewidth]{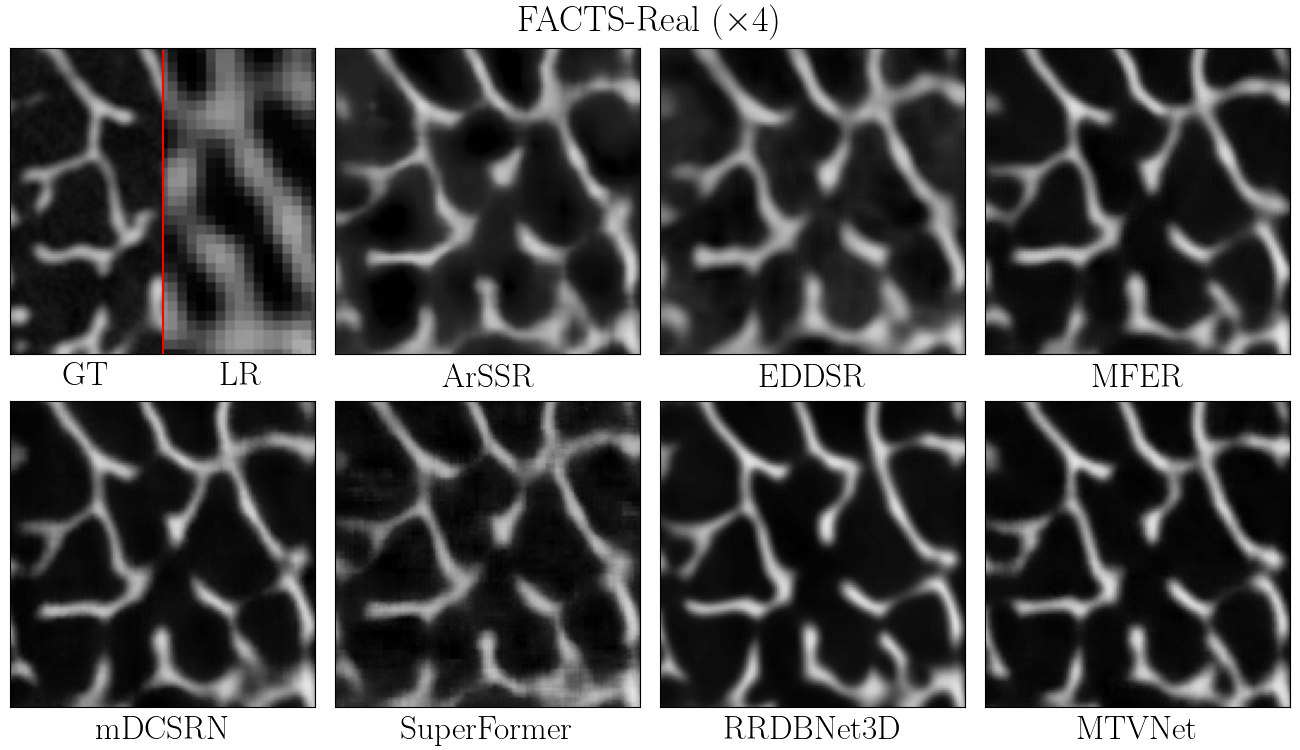}
  \end{subfigure}
  \caption{Visual comparisons of SR model outputs from the datasets HCP 1200, IXI, FACTS-Synth, and FACTS-Real using $4\times$ upscaling. The ground truth (GT) and LR input images are shown side-by-side in the top-left separated by the red line.}
  \label{fig:visual_comparison}
\end{figure*}

\subsection{Implementation details}
\label{sec:implementation_details}

All \networkname configurations use a learning rate of $2\mathrm{e}{-4}$ with no weight decay. For the brain MRI datasets (HCP 1200, IXI, BraTs 2023, and Kirby 21), we apply \networkname $L_{2}$ with two levels featuring two DCHAT blocks in the first level and three DCHAT blocks in the second level, each with six SVHAT layers. For the FACTS dataset, we use \networkname $L_{3}$ consisting of a third level with one DCHAT block on top of $L_{2}$. In all configurations of \networkname, the number of shallow features and intermediate features is set to $C_{\text{emb}} = 128$, and the number of compressed features in each skip-connection is set to $C_{\text{skip}} = 64$. In \networkname $L_{3}$ we use patch sizes $p_{3} = 8$, $p_{2} = 4$ and $p_{1} = 2$ for volumetric patch embedding at each subsequent network level while in \networkname $L_{2}$, patch sizes are set to $p_{2} = 4$ and $p_{1} = 2$. The attention window size and the size of the CAT space for each window are set to $M = 8$ and $c = 4$, respectively. To reduce memory usage during upsampling, we halve the number of features channels in \networkname, mDCSRN, SuperFormer, and RRDBNet3D before upsampling.

%% file: sec/06_results.tex
\subsection{Quantitative results}
\label{sec:results}
\cref{tab:table_quantitative_results} shows a quantitative comparison between \networkname and six other SOTA volumetric SR models ArSSR \citep{Wu_ArSSR}, EDDSR \citep{Wang_EDDSR}, MFER \citep{Li_MFER}, mDCSRN \citep{Chen_mDCSRN_C}, SuperFormer \citep{Forigua_SuperFormer}, and RRDBNet3D \citep{Wang_ESRGAN}. In the brain MRI benchmark datasets HCP 1200, IXI, BraTs 2023, and Kirby 21 our \networkname achieves second best performance across all scales. Contradicting the findings of \citet{Forigua_SuperFormer}, we observe the purely CNN-based method RDDBNet3D achieving better performance metrics than the transformer-based SuperFormer and our method. 
We reason that the advantage of RRDBNet3D may be due to local image dependencies being predominant in these datasets, limiting the benefit of the broader receptive field offered by ViTs.
Still, our \networkname achieves only slightly lower performance than RRDBNet3D while there is a greater performance gap between our method and SuperFormer, especially in $2\times$ upscaling.

In the FACTS dataset, where we can leverage the multi-contextual architecture of our proposed method, we observe several new trends: In FACTS-Synth, our proposed \networkname outperforms all other methods by a significant margin at all scales. Compared with SuperFormer, \networkname improves PSNR scores by $0.44$dB$\sim$$1.11$dB and by $0.70$dB$\sim$$1.79$dB when compared with RRDBNet3D. These improvements illustrate that additional contextual information enables significant SR performance gains in high-resolution volumetric images. In FACTS-Real where the clinical-CT images are used as LR model input, we observe CNN-based methods RRDBNet3D, MFER, and our \networkname achieve the best results. We hypothesize that this discrepancy in performance results from the domain shift between micro-CT and clinical-CT, which largely deprives the clinical-CT images of long-range image dependencies.
The trabecular structure in the clinical-CT images is largely indistinguishable, whereas the LR micro-CT images show more distinct repeatable patterns that could offer more valuable contextual information. Therefore, we surmise that the performance gains of incorporating additional long-range information in \networkname diminishes for this SR task. 

\subsection{Qualitative results}
\cref{fig:visual_comparison} shows a visual comparison of SR predictions on scale $4 \times$ for HCP 1200, IXI, FACTS-Synth, and FACTS-Real. We find that \networkname produces faithful reconstructions of structures and patterns across all datasets. Compared with ArSSR, EDDSR, MFER, mDCSRN, and SuperFormer, our \networkname produces notably sharper features while producing similar results as RRDBNet3D. In the Brain MRI datasets HCP 1200 and IXI, we find that many methods struggle to reconstruct anatomical details while RRDBNet3D and our \networkname produce the clearest results. In FACTS-Synth, we find that other models tend to produce unnaturally blurred textures, whereas our proposed \networkname suffers much less from these artifacts.

%% file: sec/07_ablation_study.tex
\subsection{Ablation experiments}
\label{sec:ablation_study}
We study the effect of our proposed features of \networkname, including the addition of CATs, shifted window hierarchical attention, and multi-contextual network levels. \cref{tab:ablation_study} shows a quantitative comparison on the BraTS 2023 dataset using $\times 4$ upscaling. Replacing the SW-MSA procedure \cite{Liu_SwinTransformer} with CAT-based hierarchical attention results in slight performance gains across all metrics, though only marginally compared to the gains seen in FasterViT \cite{Hatamizadeh_FasterViT}. Since CATs contain compressed feature summaries of each attention window, we hypothesize that this compression process discards most of the pixel-level information essential for SR. These details are less critical in image classification, hence why CATs have been observed to result in higher performance gains in this domain \cite{Hatamizadeh_FasterViT}. 
Incorporating our modified SW-MSA mechanism with CATs improves the receptive field and positively impacts performance. Finally, adding multi-contextual information in \networkname yields the largest relative improvement, increasing PSNR by $0.11$dB$\sim$$0.16$dB over other configurations. Notably, even with the relatively small volumetric samples ($\leq 240^3$ voxels) in BraTS 2023, \networkname benefits from multi-contextual information. These results highlight the value of additional contextual information, even in small-scale volumetric SR.
\begin{table}[t]
\renewcommand{\arraystretch}{1.0}

\begin{footnotesize}
\begin{tabular}{@{}ccccc@{}}
\toprule
Method        & \begin{tabular}[c]{@{}c@{}}Cyclic\\ shift\end{tabular} & CAT   & \begin{tabular}[c]{@{}c@{}}Multi\\ context\end{tabular} & \begin{tabular}[c]{@{}c@{}}PSNR/SSIM/NRMSE\\ BraTS 2023 $(\times 4)$\end{tabular} \\ \midrule
SW-MSA        & \cmark        & \xmark & \xmark                                                   & 35.00 / .9460 / .1273                                                           \\
MSA w. CAT    & \xmark        & \cmark & \xmark                                                   & 35.02 / .9463 / .1269                                                            \\
SW-MSA w. CAT & \cmark        & \cmark & \xmark                                                   & 35.05 / .9467 / .1265                                                            \\
MTVNet        & \cmark        & \cmark & \cmark                                                   & \underline{35.16} / \underline{.9477} / \underline{.1250}                                                            \\ \bottomrule
\end{tabular}
\end{footnotesize}
\caption{Ablation on the proposed features of \networkname. The best performance metrics PSNR/SSIM/NRMSE are underlined.}
\label{tab:ablation_study}
\end{table}

%% file: sec/08_conclusion.tex
\subsection{Memory footprint of \networkname}

\begin{figure}[t]
    \centering
    \includegraphics[width=1.0\linewidth]{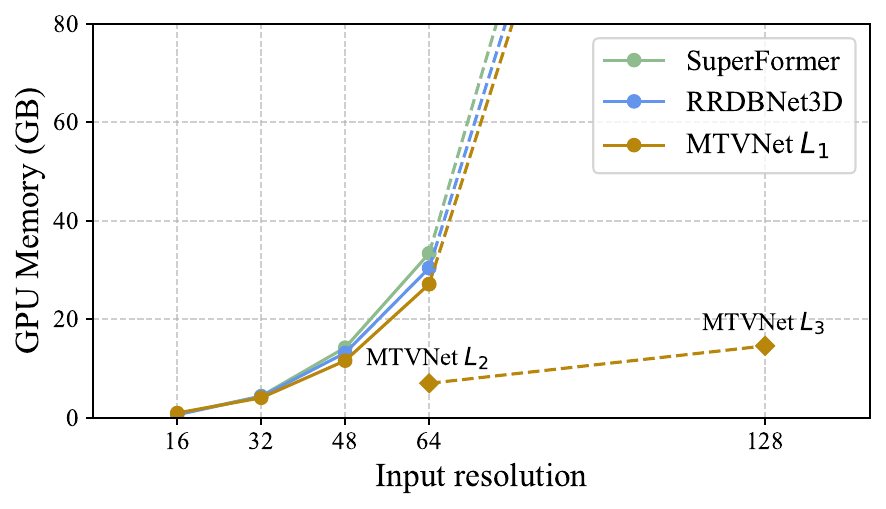}
    \caption{GPU Memory usage of SuperFormer, RRDBNet3D, and \networkname using a single 3D patch at resolutions $16^3$, $32^3$, $48^3$, and $64^3$. Adding contextual levels to \networkname enables increasing resolution to $128^3$ and beyond without exceeding GPU memory.}
    \label{fig:memory_vs_resolution}
\end{figure}

\cref{fig:memory_vs_resolution} shows the memory footprint required by SuperFormer, RRDBNet3D, and \networkname using different volumetric input resolutions. Using a single level, \networkname $L_{1}$ requires less memory than SuperFormer and RRDBNet3D. Furthermore, provided the prediction area is fixed to $32^3$, adding more network levels to \networkname allows processing volumetric input sizes far exceeding the capabilities of other volumetric SR architectures.

\section{Conclusion}
\label{sec:conclusion}
In this work, we present \networkname, a transformer-based approach for volumetric SR tailored for high-resolution 3D data. To overcome the challenge of limited contextual information in volumetric SR, we propose a multi-contextual network structure with a coarse-to-fine feature extraction and tokenization scheme. This approach reduces the number of tokens needed to cover large volumetric regions, allowing our model to process significantly larger input sizes than competing methods.
To enhance long-range information exchange in the expanded input volume, we implement a novel shifting volumetric hierarchical attention transformer (SVHAT) layer inspired by FasterViT \cite{Hatamizadeh_FasterViT} and SwinV2 \cite{Liu_SwinV2} that employs a combination of full and window-based attention to capture both global and local image dependencies. We evaluate the performance of \networkname against other volumetric SR approaches across several data domains, including brain MRI data and high-resolution CT data. Based on extensive experiments, we make several conclusions: In contradiction with the current research trends in 2D SR, we observe CNN-based models outperform transformer-based models in certain data domains. The effectiveness of CNN-based SR models is especially pronounced in lower-resolution 3D samples where the larger receptive field of transformers cannot be leveraged as effectively. Nevertheless, our proposed \networkname with extra contextual processing layers outperforms all other models given high-resolution 3D data with meaningful long-range image dependencies. 

We surmise that our multi-contextual approach for volumetric image processing could be greatly beneficial for other vision applications such as segmentation, classification, and recognition in volumetric images. 


\newpage

%% file: sec/A1_suppl.tex
\clearpage
\setcounter{page}{1}
\maketitlesupplementary

\section{Details of SVHAT layer}

\begin{figure}[h]
   \centering
   \includegraphics[width=0.8\linewidth]{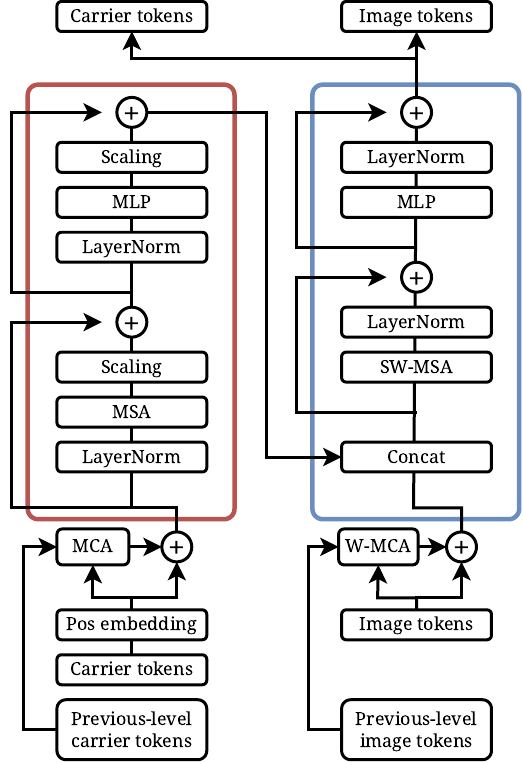}
   \caption{Overview of SVHAT featuring attention branches for CATs (\textcolor{red}{red}) and ITEs (\textcolor{blue}{blue}). We use MCA and W-MCA to merge tokens from previous network levels before computing attention in each branch.}
   \label{fig:svhat_layer}
\end{figure}

\cref{fig:svhat_layer} provides an overview of our proposed shifting volumetric hierarchical attention transformer (SVHAT) layer. Our SVHAT captures global and local token dependencies using separate attention branches for CATs and ITEs. The CAT attention branch (shown in red) follows the attention procedure of FasterViT \cite{Hatamizadeh_FasterViT}, whereas the ITE branch (shown in blue) follows the approach of SwinV2 \cite{Liu_SwinV2}. 
Before computing attention in each branch, SVHAT uses full multi-head cross-attention (MCA) for computing attention between CATs extracted from different network levels in \networkname. Similarly, windowed multi-head cross-attention (W-MCA) is used for computing attention between ITEs from different network levels.    

\section{Datasets}

\textbf{Human Connectome Project (HCP) 1200}\\
The HCP 1200 Subjects Data Release \cite{Van_HCP_1200_dataset} includes structural MRI scans from 1113 healthy subjects acquired using a 3T scanning platform. We use the T1-weighted images which feature an isotropic resolution of 0.7 mm and a volume size of $320\times320\times256$ voxels. Following the approach in \cite{Chen_mDCSRN_C, Forigua_SuperFormer}, the dataset is split into 780 subjects for training, and 111 subjects each for validation, evaluation, and testing. Performance evaluation is performed using the 111 subjects in the test set.  

\noindent
\textbf{Information eXtraction from Images (IXI)}\\
The IXI dataset contains multi-modality MRI data (PD-, T1- and T2-weighted images) collected from a total of 600 healthy subjects scanned using one 3T, and two 1.5T platforms. We use all 581 T1-weighted images of IXI, of which 507 scans feature a resolution of $0.9375\times0.9375\times1.2$~mm and a volume size of $256\times256\times150$ voxels, and the remaining 74 scans feature a resolution of $0.93749\times0.9375\times1.2$ mm and a volume size of $256\times256\times146$ voxels.  The dataset is split into 500 subjects for training, 6 for validation, and 75 for testing. Performance evaluation is performed using the 75 subjects in the test set.  

\noindent
\textbf{Brain Tumor Segmentation Challenge (BraTS) 2023}\\
For BraTS 2023, we use the T1-weighted structural MRI images of the Adult Glioma segmentation challenge \cite{Baid_BraTS_GLI, Menze_BraTS_GLI, Bakas_BraTS_GLI, Bakas_BraTS2023_GLI_2}. This subset contains a total of 1,470 scans collected from glioma patients. The images are skull-stripped and standardized to an isotropic resolution of 1~mm and a volume size of $240\times240\times155$ voxels. We use the dataset split provided by the Adult Glioma segmentation challenge, which allocates 1,251 subjects for training and 219 for validation. Performance evaluation is performed using the 219 subjects in the validation set.

\noindent
\textbf{Kirby 21}\\
The Kirby 21 dataset \cite{Landman_Kirby21_dataset} includes multi-modality MRI images acquired from healthy individuals with no history of neurological conditions. We use all 42 T2-weighted images of Kirby 21 which feature a resolution of $1\times0.9375\times0.9375$ mm and a volume size of $180\times256\times256$ voxels. The dataset is split into 37 images for training (KKI-06 to KKI-42) and 5 for testing (KKI-01 to KKI-5). Performance evaluation is performed using the 5 subjects in the test set.  

\noindent
\textbf{Femur Archaeological CT Superresolution (FACTS)}\\
The FACTS dataset consists of 12 archaeological proximal femurs scanned using clinical-CT and micro-CT platforms \cite{Bardenfleth_FACTS}. The clinical-CT and micro-CT scans feature a resolution of $0.21\times0.21\times 0.4$ mm and $58\times58\times58$ $\mu$m, respectively. The clinical-CT volumes are registered and linearly interpolated to match the volume sizes of the micro-CT images. The dataset is split into 10 images for training and 2 images (f\_002 and f\_138) for testing and subsequent performance evaluation.

\section{Visual comparisons using LAM}

\begin{figure}[t]
    \centering
        \begin{subfigure}{\linewidth}
        \includegraphics[width=1.0\linewidth]{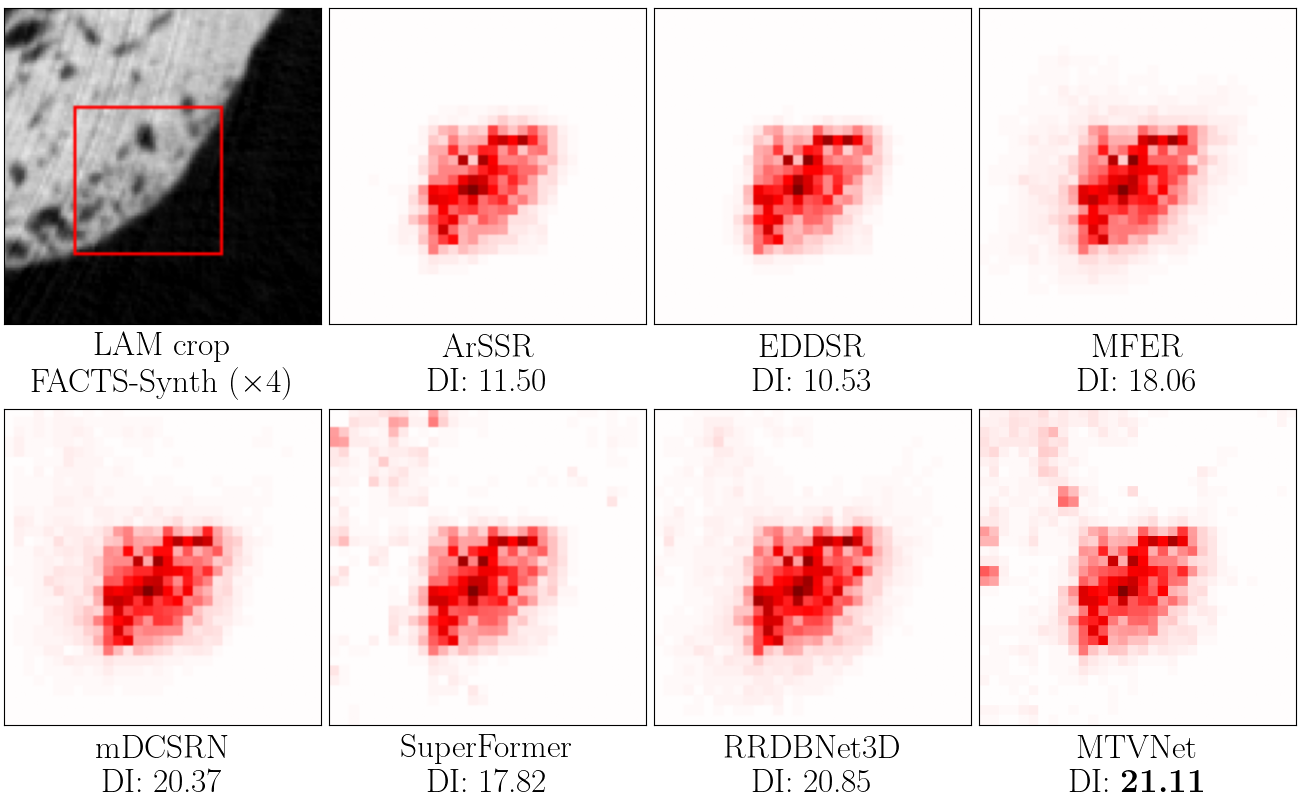}
        \end{subfigure}
        \begin{subfigure}{\linewidth}
        \includegraphics[width=1.0\linewidth]{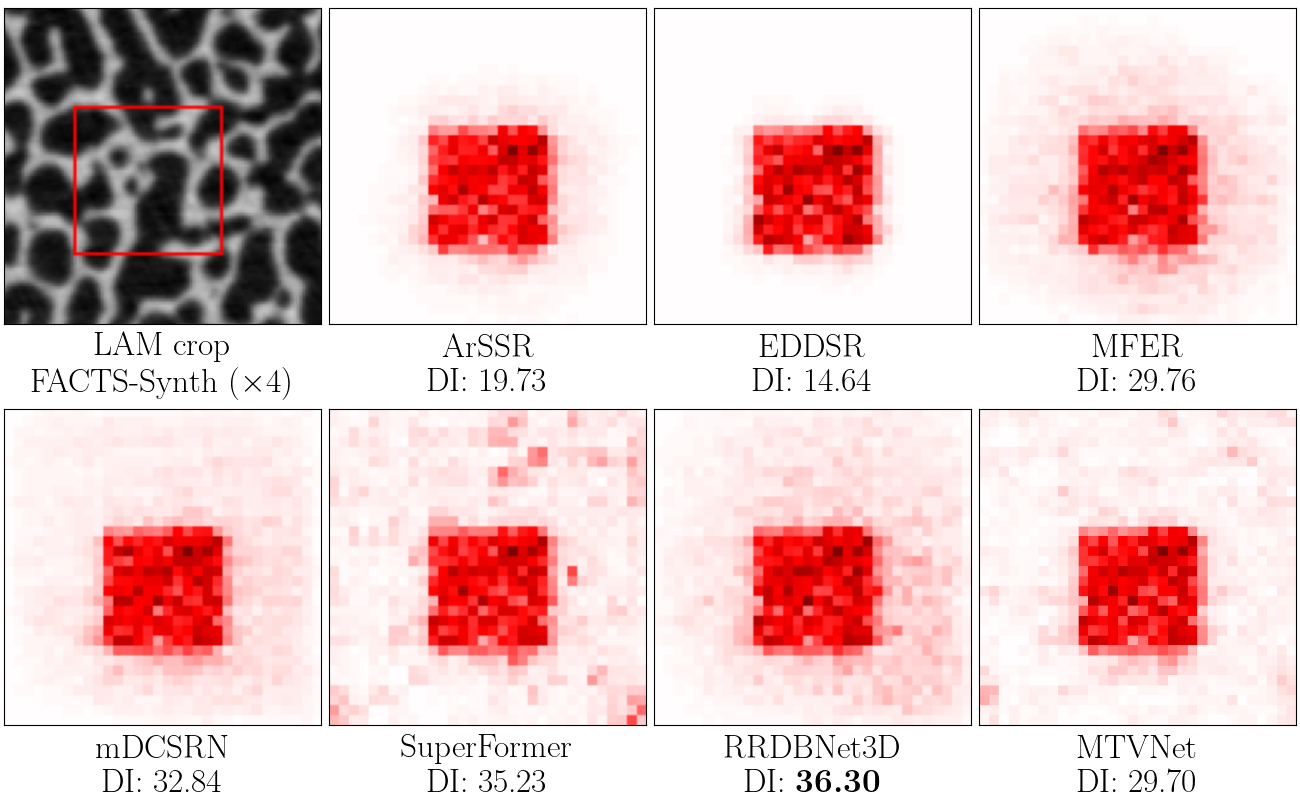}
        \end{subfigure}
        \begin{subfigure}{\linewidth}
        \includegraphics[width=1.0\linewidth]{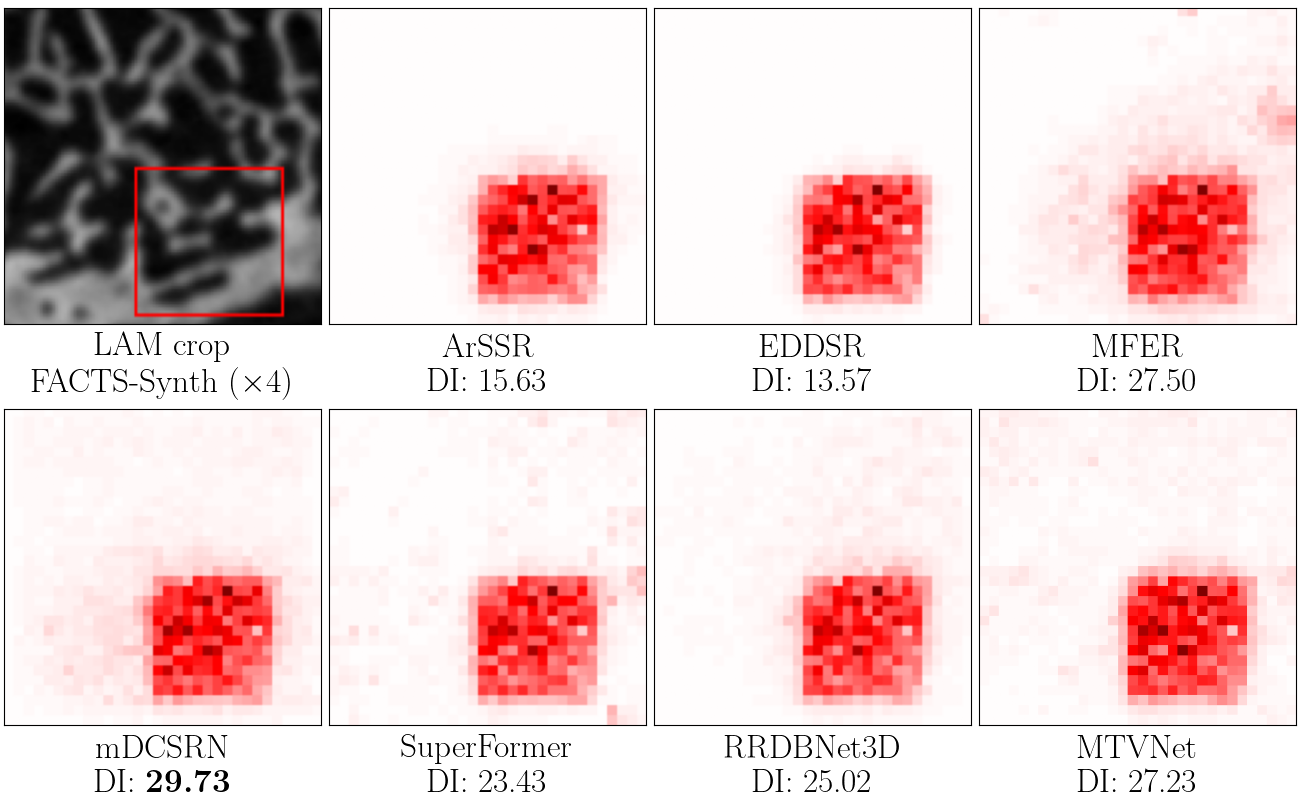}
        \end{subfigure}
        \caption{LAM comparisons of SR models using FACTS-Synth at $\times 4$ upscaling. Each LAM image shows the average contribution of each voxel throughout all slices of the prediction volume marked by the red box.}
        \label{fig:LAM_comparisons_supp}
\end{figure}

We investigate how effectively volumetric SR models utilize the surrounding image context when computing SR predictions. To this end, we employ the LAM attribution method \citep{gu2021interpreting}, which is a modification of the integrated gradient method \citep{sundararajan2017axiomatic} designed to investigate SR architectures. We extend the LAM framework for volumetric SR and visualize the range of involved input voxels for all volumetric SR models. \cref{fig:LAM_comparisons_supp} shows a visual comparison of the LAM results using three sample volumes from the FACTS-Synth dataset at $\times4$ upscaling. To visualize the contribution of each voxel, each LAM image shows the average contribution of each voxel throughout all slices of the prediction volume. The red regions highlight the input voxels contributing to the SR prediction volume marked by the red box, with higher intensities indicating a stronger voxel influence on the prediction output. We also report the diffusion index (DI), which is a measure of the overall range of involved voxels used to predict the SR output.

In contrast to LAM results reported in 2D SR \cite{Chen_Activating, Hsu_DRCT, Chu_HMANet}, we find that there is a very sharp decline in contribution from input voxels outside the prediction volume for all models across all three sample volumes. This trend suggests that even in high-resolution datasets such as FACTS-Synth, local information is of relatively higher importance for volumetric SR than for 2D SR. Furthermore, we find that the degree to which the surrounding input voxels contribute to the SR prediction is highly dependent on the image structure inside the sample volume. 

Our analysis finds no consistent top-performing model in terms of DI across the considered sample volumes. Given our experimental results in \cref{tab:table_quantitative_results}, we find no strong evidence correlating higher DI to higher PSNR/SSIM/NRMSE scores. Notably, we observe that for convolution models, the contribution of distant input voxels contribute progressively less to the SR output, whereas the LAM result of the transformer-based models SuperFormer and \networkname reveals areas of high contribution far outside the prediction volume.

\section{More visual comparisons}

\begin{figure*}[h]
  \centering
  \begin{subfigure}{0.49\linewidth}
    \includegraphics[width=1.0\linewidth]{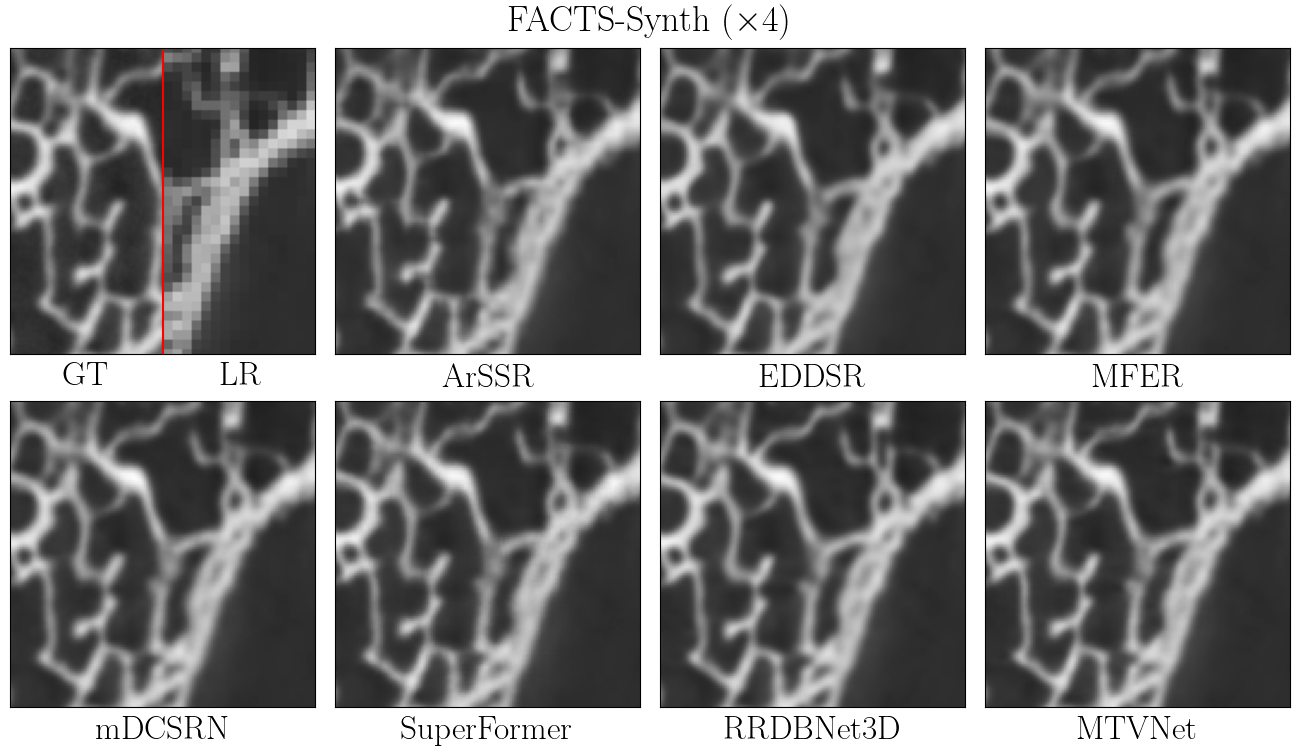}
  \end{subfigure}
  \hfill
  \begin{subfigure}{0.49\linewidth}
    \includegraphics[width=1.0\linewidth]{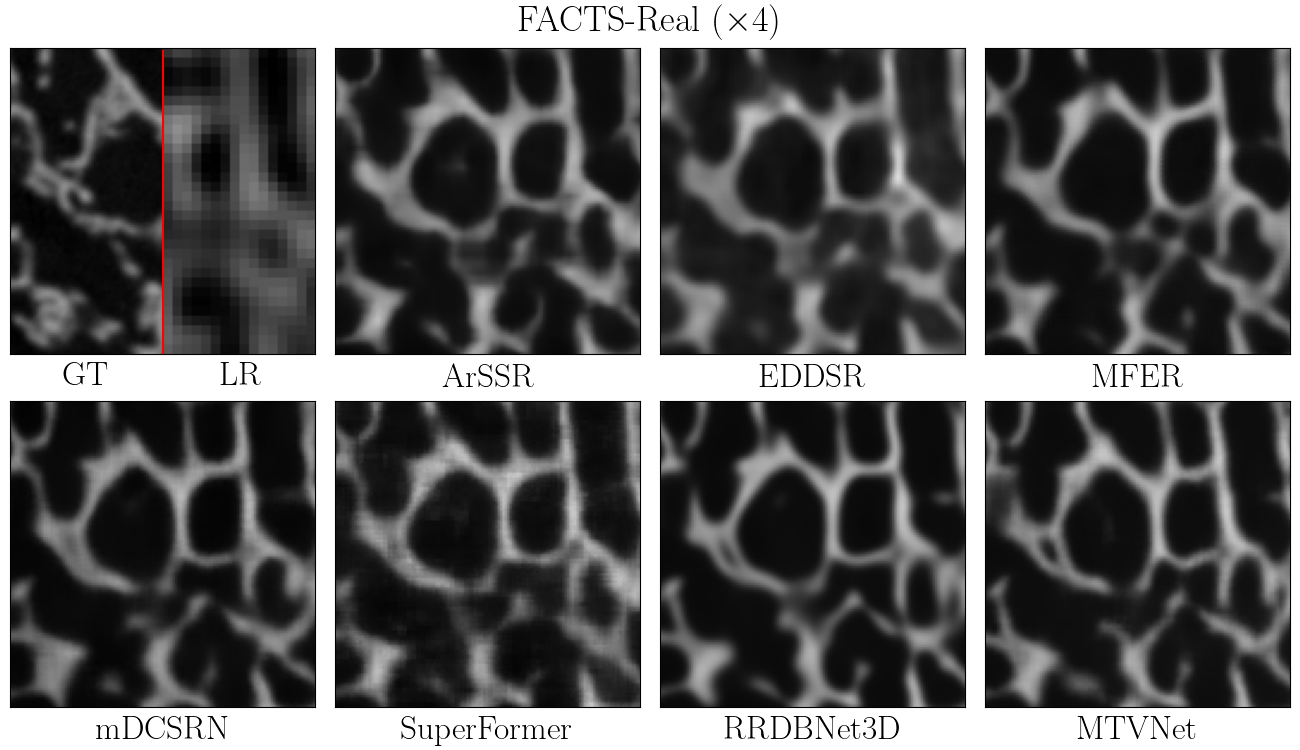}
  \end{subfigure}
  
  \vspace{0.5em} 

  \begin{subfigure}{0.49\linewidth}
    \includegraphics[width=1.0\linewidth]{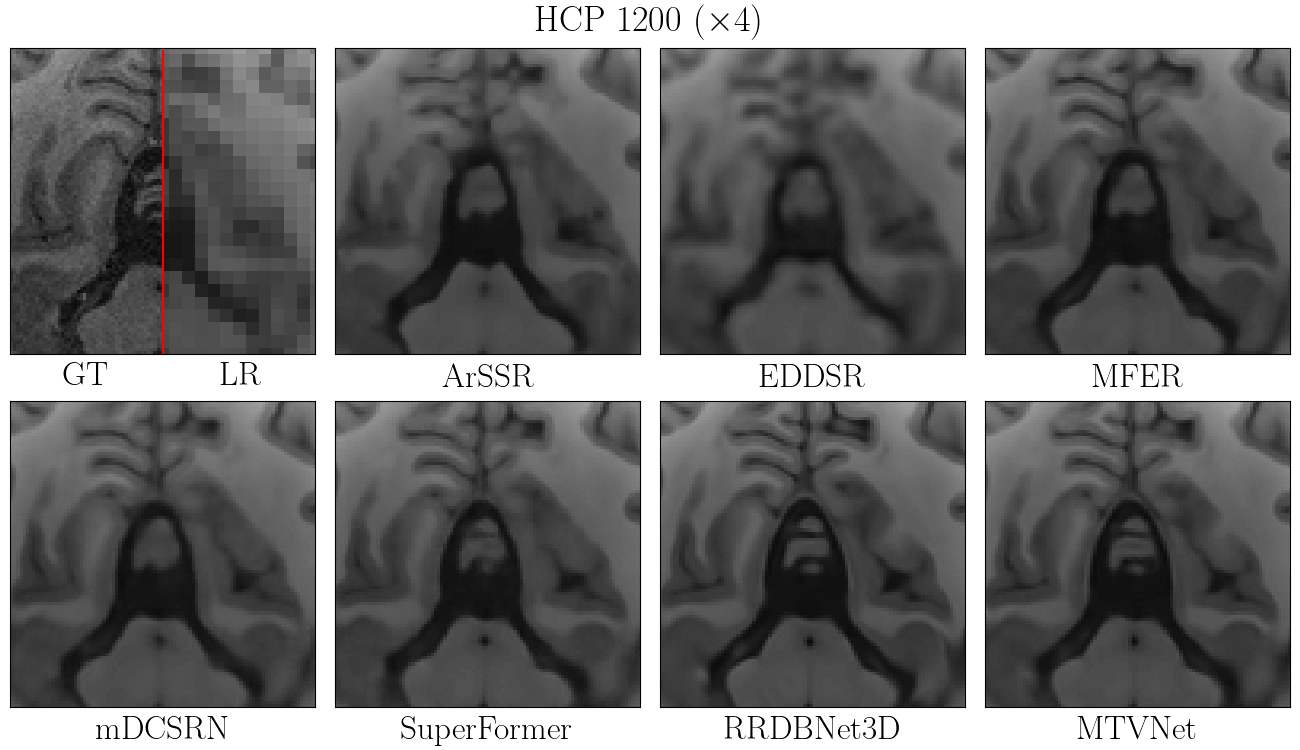}
  \end{subfigure}
  \hfill
  \begin{subfigure}{0.49\linewidth}
    \includegraphics[width=1.0\linewidth]{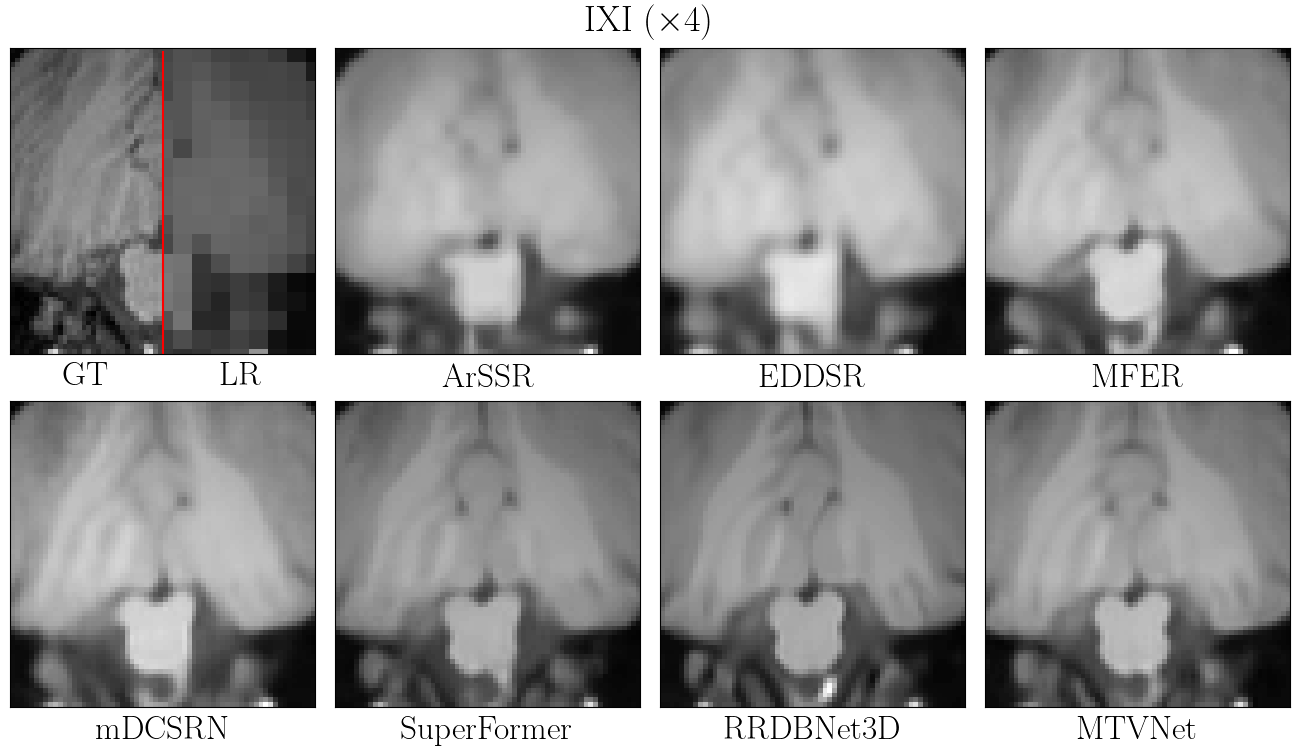}
  \end{subfigure}

  \vspace{0.5em} 

  \begin{subfigure}{0.49\linewidth}
    \includegraphics[width=1.0\linewidth]{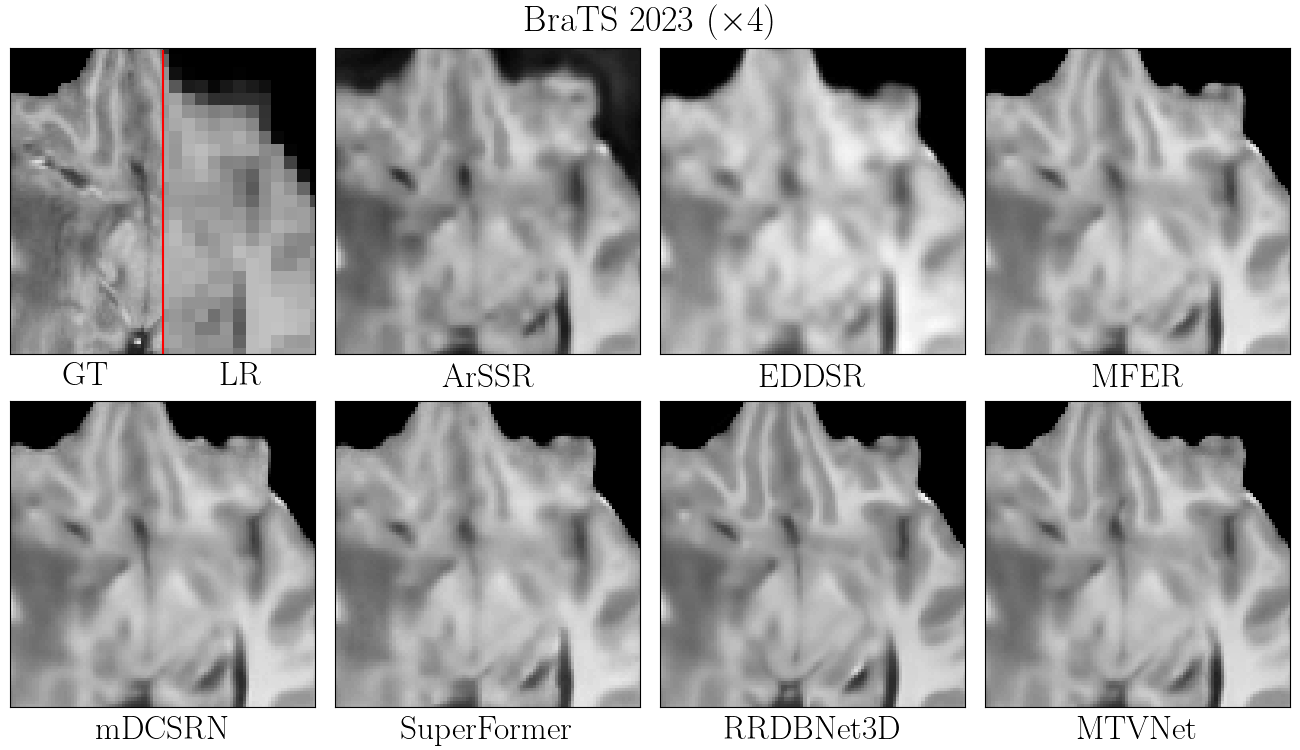}
  \end{subfigure}
  \hfill
  \begin{subfigure}{0.49\linewidth}
    \includegraphics[width=1.0\linewidth]{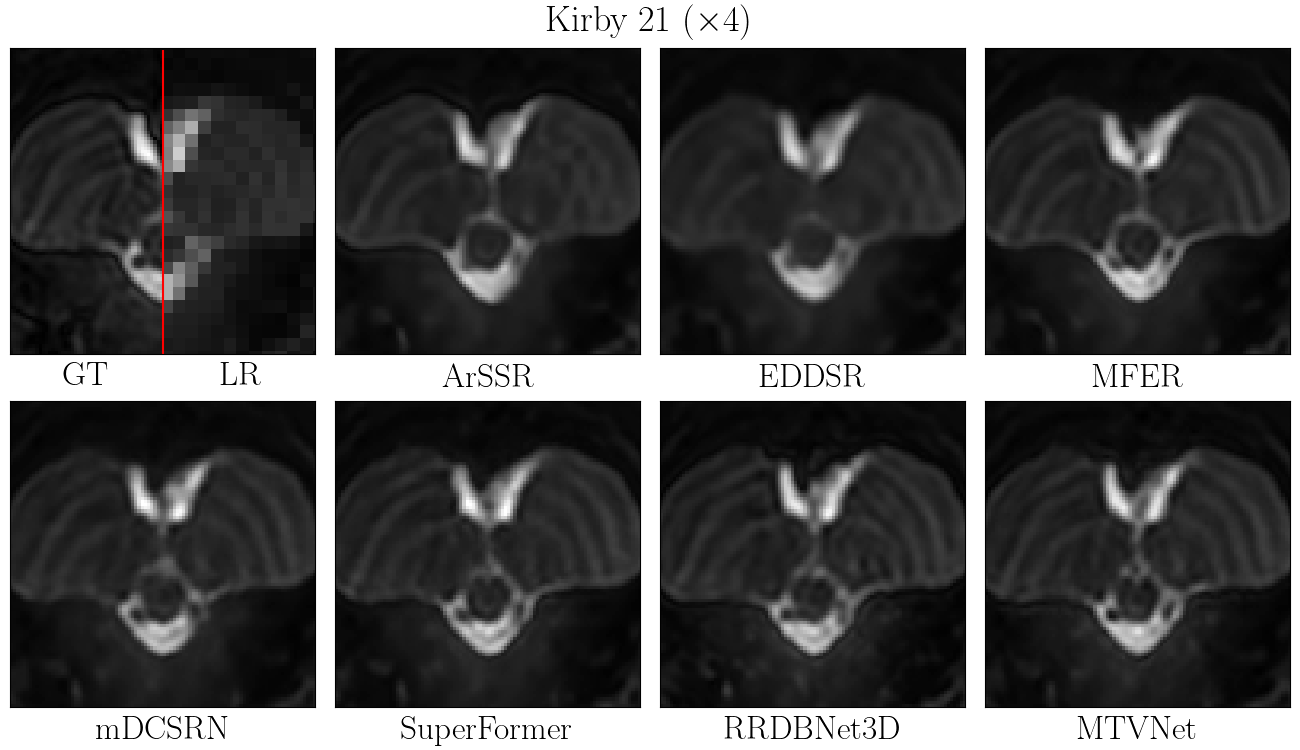}
  \end{subfigure}
  \caption{Visual comparisons of SR model outputs from the datasets HCP 1200, IXI, BraTS 2023, Kirby 21, FACTS-Synth, and FACTS-Real using $4\times$ upscaling. The ground truth (GT) and LR input images are shown side-by-side in the top-left separated by the red line.}
  \label{fig:visual_comparison_supp}
\end{figure*}

\cref{fig:visual_comparison_supp} shows more visual comparisons of SR predictions using the datasets HCP 1200, IXI, BraTS 2023, Kirby 21, FACTS-Synth and FACTS-Real at $\times 4$ upscaling. Across the four structural brain MRI datasets, our \networkname produces noticeably sharper edges than ArSSR, EDDSR, MFER, mDCSRN, and SuperFormer while performing on par with RRDBNet3D.  
In FACTS-Real, the other transformer-based model SuperFormer reproduces texture artifacts not seen in the ground truth image. Our \networkname avoids these artifacts while producing a less blurry SR prediction.   

